\documentclass[]{files/interact}
\usepackage{hyperref,algorithmic,algorithm,tikz,pgfplots,pgfplotstable, siunitx, comment}
\usepackage{epstopdf}
\usepackage[caption=false]{files/subfig}
\usepackage[numbers,sort&compress]{files/natbib}
\bibpunct[, ]{[}{]}{,}{n}{,}{,}
\renewcommand\bibfont{\fontsize{10}{12}\selectfont}
\makeatletter
\def\NAT@def@citea{\def\@citea{\NAT@separator}}
\makeatother
\theoremstyle{plain}

\theoremstyle{definition}

\theoremstyle{remark}

\begin{document}

\title{ \Large{Enhancing Aerial Combat Tactics through \newline Hierarchical Multi-Agent Reinforcement Learning}}

\author{\name{Ardian Selmonaj\textsuperscript{a}, Oleg Szehr\textsuperscript{a}, Giacomo Del Rio\textsuperscript{a}, Alessandro Antonucci\textsuperscript{a}, \newline 
Adrian Schneider\textsuperscript{b} and Michael R\"{u}egsegger\textsuperscript{b}}
\affil{\textsuperscript{a}Istituto Dalle Molle di Studi sull'Intelligenza Artificiale (IDSIA), USI-SUPSI, Switzerland; \textsuperscript{b}Armasuisse Science + Technology, Switzerland}}
\maketitle

\begin{abstract}
This work presents a \emph{Hierarchical Multi-Agent Reinforcement Learning} framework for analyzing simulated air combat scenarios involving \emph{heterogeneous} agents. The objective is to identify effective \emph{Courses of Action} that lead to mission success within preset simulations, thereby enabling the exploration of real-world defense scenarios at low cost and in a safe-to-fail setting. Applying deep Reinforcement Learning in this context poses specific challenges, such as complex flight dynamics, the exponential size of the state and action spaces in multi-agent systems, and the capability to integrate real-time control of individual units with look-ahead planning. To address these challenges, the decision-making process is split into two levels of abstraction: low-level policies control individual units, while a high-level commander policy issues macro commands aligned with the overall mission targets. This hierarchical structure facilitates the training process by exploiting policy symmetries of individual agents and by separating control from command tasks. The low-level policies are trained for individual combat control in a \emph{curriculum} of increasing complexity. The high-level commander is then trained on mission targets given pre-trained control policies. The empirical validation confirms the advantages of the proposed framework.
\end{abstract}


\section{Introduction}\label{sec:intro}
We investigate deep \emph{Reinforcement Leaning} (RL) as a method for the exploration of preset air combat scenarios at a low cost and in a safe-to-fail setting. This is motivated by the strong performance of RL agents in finding effective \emph{Courses of Action} (CoA) across a wide range of environments, including combinatorial settings such as Chess or Go~\cite{chess_go}, real-time continuous control tasks found in arcade video games~\cite{mnih2013playing}, and scenarios that combine control with strategic decision-making, as seen in modern wargames~\cite{hhmarl2d}. The application of RL in the context of air combat comes with a number of specific challenges. Those include structural properties of the simulation scenario, such as the complexity of the individual units and their flight dynamics, the exponential size of the combined state and action spaces, the depth of the planning horizon, the presence of stochasticity and imperfect information, etc. Overall the size of the game tree (i.e.,~the set of possible CoAs) in strategic games and defense scenarios appears vast and beyond the access of straightforward search. Furthermore, real-world operations involve the simultaneous maneuverings of individual units, but also being mindful of the strategic positions and global mission planning. Training policies that integrate real-time control at the troop level with high-level mission planning at the commander level is challenging, as these tasks inherently demand distinct system requirements, algorithmic approaches, and training configurations.

To address these challenges and replicate real-world defense operations, this work\footnote{This article is a revised and extended version of the material originally published as a conference paper~\cite{hhmarl2d}.} investigates a hierarchical \emph{Multi-Agent Reinforcement Learning} (MARL) framework for analyzing simulated air combat scenarios involving heterogeneous agents. Our approach decomposes the decision-making process into two levels of abstraction where \emph{low-level} policies govern the real-time control of individual units, while a \emph{high-level} policy issues macro commands aligned with the overall mission targets. 
The low-level policies are trained on a list of preset scenarios (such as attack/evade) with a command flag that identifies the scenario target. To enhance robustness and learning efficiency, their training is organized in a learning \emph{curriculum} with increasingly complex training scenarios and league-based self-play. 
The high-level policy learns to assign appropriate flags to subordinate agents based on evolving mission goals. Decision authority is delegated to the higher-level commander for strategic planning, while low-level actors autonomously execute individual control tasks. Just as in real-world operations, this structure greatly facilitates the training process of policies, by exploiting policy symmetries of low-level actors and by targeting the flow of information to the relevant decision authority. This architecture also supports a clear separation between control and command, allowing task-specific policies to be developed using specialized training schemes.

In summary, we propose a hierarchical MARL framework to address decision-making in complex air combat simulations by integrating simultaneous maneuvering with tactical mission planning. Our key contributions are as follows:

\begin{enumerate}
    \item We develop a lightweight environment platform suitable for fast simulation of core agent dynamics and interactions. Despite restricting motion to 2D by assuming constant aircraft altitude, the system realistically captures agent interactions and maneuvering behavior.
    \item Our training setup employs fictitious self-play through curriculum learning with increasing levels of complexity to improve combat performance.
    \item We design sophisticated neural network architectures incorporating attention mechanisms, recurrent units and parameter sharing to train both low-level control policies and a high-level commander policy.
    \item Since deep learning systems often exhibit black-box characteristics and pose risks in scientific evaluation, our approach addresses these issues by enabling explainability through the analysis of its hierarchical components. 
\end{enumerate}

\subsection{Outline}\label{sec:outline}
Sec.~\ref{sec:related} summarizes the state-of-the-art and describes how our work extends on the existing literature. Sec.~\ref{sec:overall_concept} introduces the base features of the aircraft simulator and principles of MARL. Sec.~\ref{sec:method} presents air engagement scenarios and describes our training procedure. Our experimental findings are presented in Sec.~\ref{sec:experiments}, while our conclusions and possible future works are discussed in Sec.~\ref{sec:conclusions}.

\section{Related Work}\label{sec:related}

The literature in air combat scenarios incorporating RL methods can roughly be divided into three categories: small engagements (single-agent), large engagements (multi-agent) and hierarchical methods.

\subsection{Single-Agent Techniques}
Aerial combat tactics have been studied in the simulation and modeling literature for decades, with a significant focus on 1-vs-1 scenarios. Research on small engagements typically addresses unit \emph{control}, i.e., how a combat unit should maneuver to achieve a favorable outcome. This often involves gaining a positional advantage that enables firing with minimal exposure to return fire~\cite{shawFight}. Common approaches include expert systems~\cite{burgin_1,burgin_2,jones_1,jones_2}, control laws for pursuit/evasion~\cite{Eklund,Virtanen2004,Virtanen2006,You2014}, game-theoretic models~\cite{Jarmark1984,Merz1985,Greenwood1992}, machine learning~\cite{McGrew2010,Ma2018,Day2018,Vlahov2018}, and hybrid systems~\cite{Smith2000,Smith2000_,Smith2001,Smith2002,Smith2004,Toubman2015,Toubman2016}. RL gained traction in this context as deep RL systems (i.e., combining RL and deep neural networks) achieved superhuman performance across a range of control-heavy games, from Atari environments~\cite{mnih2013playing}, to strategic games like Go and Chess~\cite{chess_go}, and combined real-time control and strategy games such as Starcraft 2~\cite{startcraft2} and Dota 2~\cite{dota2}. RL systems offer an objective and flexible framework for deriving strong CoAs, unrestricted by the quantity or quality of expert data. Given the success of various RL models across game domains, it is natural to explore their applicability in defense modeling. For example, DQN~\cite{mnih2013playing}, which has demonstrated strong performance in Atari games, has been adapted and applied to air combat scenarios~\cite{rl_combat6}. Other works apply \emph{Deep Deterministic Policy Gradient} (DDPG)~\cite{rl_combat0, rl_combat1} or A3C~\cite{rl_combat7} for \emph{Unmanned Aerial Vehicle} (UAV) maneuver learning. Curriculum-based approaches with gradually increasing complexity are explored in~\cite{rl_combat3,rl_combat4}. Finally, league play serves as a critical mechanism to prevent overfitting to specific opponent strategies in self-play settings, where cyclic overfitting can degrade overall performance~\cite{startcraft2,league_play,rl_combat5}.

\subsection{Multi-Agent Techniques}

In multi-agent scenarios, the number of possible combinations of agent states and actions increases exponentially with the number of agents. For RL systems, this poses a significant challenge as the exploration of such spaces can be beyond the access of computational resources. MARL systems make an important step towards the resolution of this 'curse of dimensionality' by exploiting symmetries within individual agents. If, for example, agents are interchangeable for a certain task, they could be equipped with identical policies, which greatly facilitates the overall training and execution processes~\cite{Gronauer2021}. In defense modeling, research in multi-unit engagements frequently focuses on weapon-target assignment~\cite{optimumTargetAssignment}, pilot-like decision-making ~\cite{Tidhar1998}, and high-level tactical decisions~\cite{dayThesis}, i.e., on \emph{planning} of CoAs rather than the control of individual units. A maneuvering strategy for UAV swarms is developed in~\cite{marl2}, but the discussion is limited to one-to-one or multi-to-one combat. The articles~\cite{marl4} and~\cite{marl5} use attention-based neural networks in the context of air combat. More advanced approaches~\cite{marl_combat_new1, marl_combat_new2} operate in 3D dynamics and introduce adaption to the popular \emph{Proximal Policy Optimization} (PPO) algorithm~\cite{ppo}. Other works incorporate heterogeneous agent dynamics~\cite{marl_combat_new3} or train UAV control maneuvers while ensuring explainable policies~\cite{marl_combat_new4}, while other approaches enhance the mission strategies through tactical reward shaping~\cite{marl_combat_new5} or incorporate communication during training~\cite{marl_combat_new6}. A focus on ad-hoc cooperation with unforeseen agents in large-scale engagements by using graph structures is presented in~\cite{marl_combat_new7}. A compact review on articles based on recent approaches for air combat modeling using RL methods can be found in~\cite{aircombat_survey}. 

\subsection{Hierarchical Techniques}
There appears to be relatively little research in defense modeling that combines \emph{Hierarchical Reinforcement Learning} (HRL) and MARL. Combining HRL within MARL could facilitate the adoption of advanced strategies by structuring decision-making and enhancing tactical complexity and effectiveness. A Hierarchical MARL approach is proposed in~\cite{marl3} to train agents using QMIX~\cite{rashid2018qmix}. An approach similar to the one presented in this paper, focusing on heterogeneous agents of two types, was explored in~\cite{marl_hieR_hetero}. The high-level target allocation agents are trained using DQN, and the low-level cooperative attacking agents are based on PPO. However, they follow the goal of \emph{Suppression of Enemy Air Defense} (SEAD). A recent approach using hierarchical MARL approach for many-to-many defense operations for \emph{Unmanned Underwater Vehicles}~(UUV) is introduced in~\cite{hmarl_combat_new8}. However, unlike the concept of SEAD or the dynamics of UUV, dogfighting aims at defeating enemy aircraft in direct air combat.

This article investigates air combat scenarios by focusing on coordinated dogfighting strategies with heterogeneous agents, employing hierarchical MARL for policy optimization. By utilizing a cascaded league-play training scheme, this work introduces structured levels of complexity that closely mimic real-world dynamics. The inclusion of hierarchical structures in the MARL framework allows for differentiated roles and decision-making among agents. The proposed approach not only aims at improving the tactical effectiveness but also contributes to our understanding of advanced cooperative strategies in highly dynamic environments.

\section{Foundational Concepts}\label{sec:overall_concept}

In this section, we describe the properties of our aircraft simulator and the relevant concepts in RL that are involved in our approach.

\subsection{Aircraft Dynamics}\label{sec:aircraft_dynamics}
We base our modeling on the dynamics of the \emph{Dassault Rafale} fighter aircraft.\footnote{ \href{https://www.dassault-aviation.com/en/defense/rafale}{https://dassault-aviation.com/en/defense/rafale}.} We focus on hierarchical coordination of multiple heterogeneous agents in 2D (assuming a constant altitude of our aircraft). While reducing the complexity of analysis by omitting the dynamics associated with the third dimension, a 2D model still retains the fundamental characteristics of air combat scenarios. By focusing on two dimensions, our approach simplifies calculations and visualizations, making it easier to model and analyze key interactions and strategies. Despite its simplicity, a 2D model can provide valuable insights into the critical aspects of aerial engagements, such as positioning, maneuvering, and timing, which are essential for studying and simulating air combat tactics. This simplification allows to focus on core combat dynamics without the computational overhead and complexity introduced by a three-dimensional model.

In air combat, engagements can be classified into two main categories: \emph{Beyond Visual Range} (BVR) and \emph{Within Visual Range} (WVR) scenarios~\cite{wvr}. BVR involves combat situations where adversaries engage at distances that exceed the line of sight, relying on long-range sensors and weapons. In contrast, WVR scenarios occur when opponents are close enough to be visible to each other. Often shorter distances necessitate maneuvering at higher frequency. Additionally, they are characterized by the presence of more detailed adversary information, while larger distances might entail imperfect information scenarios.
In this article, our focus lies on WVR combat in a perfect information setting. This focus allows us to delve into the details of maneuvering, weapon selection, and tactical decisions that define the outcome of such encounters.

Given that real-world combat scenarios often involve a variety of aircraft types, each with unique capabilities and dynamics, our approach concentrates on two specific types of aircraft that exhibit different flight characteristics. By selecting these two models, we introduce a reasonable level of \emph{heterogeneity} into our simulations, mirroring the diversity typically encountered in actual aerial engagements. Our model allows to explore the strategic interactions and outcomes that might arise from the varied capabilities and behaviors of different aircraft types in air combat. 

\begin{figure}[htb]
\centerline{\includegraphics[scale=0.13]{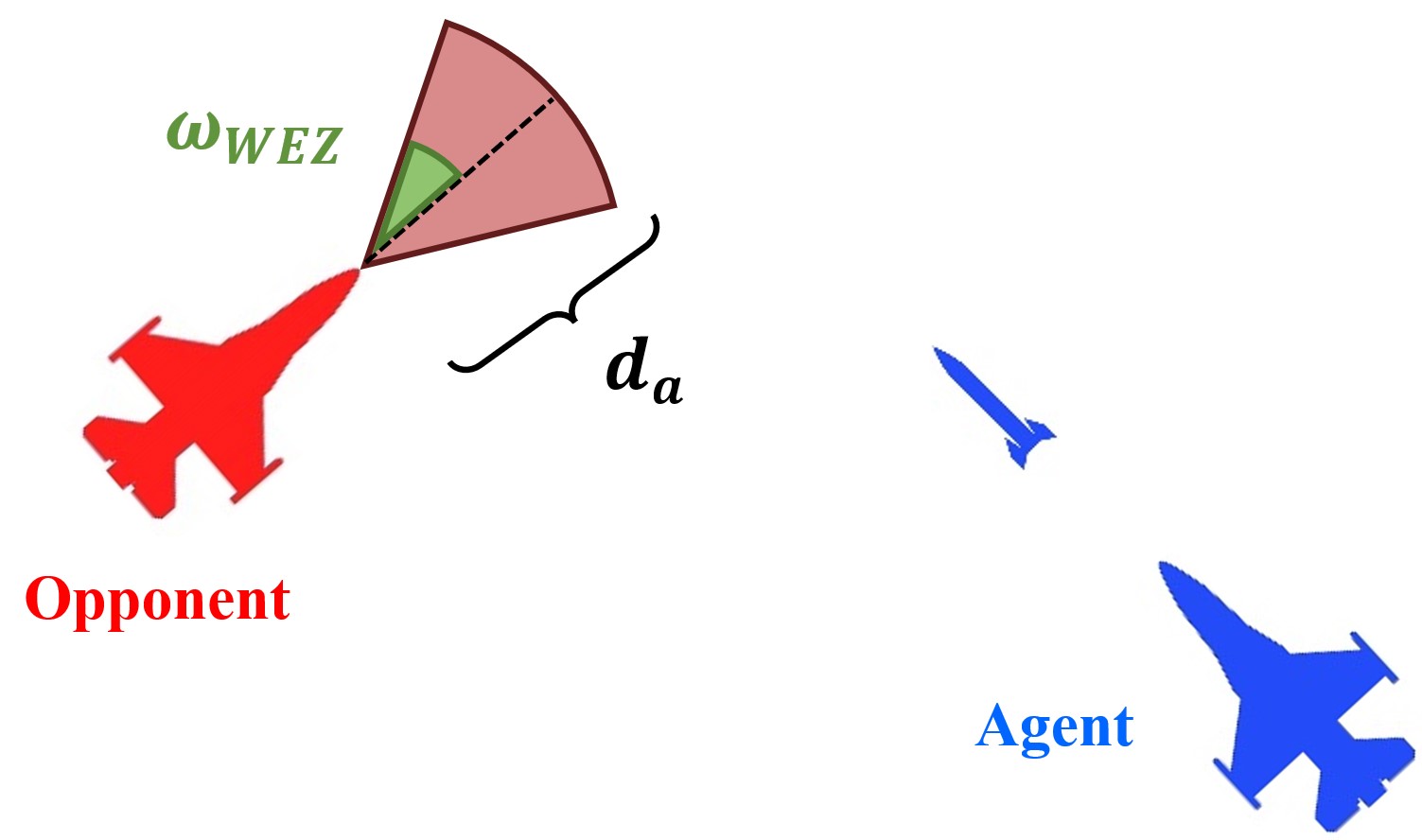}}
\caption{Aircraft attacking mechanism. Blue aircraft are (our) RL agents, red aircraft denote opponents. Shooting with a cannon is represented by a conical shape (WEZ), while firing a rocket is depicted with a corresponding symbol.}
\label{fig:aircrafts_attack}
\end{figure}

Therefore, the first aircraft (\textbf{AC1}) is more agile and equipped with rockets, while the second type (\textbf{AC2}) has no rockets but longer cannon range. The dynamics of AC1 and AC2 are characterized as shown in Table~\ref{table:aircraft_dynamics}, while Fig.~\ref{fig:aircrafts_attack} illustrates the attacking mechanisms. Cannon shots are modeled by a conical shaped area, called \emph{Weapon Engagement Zone} (WEZ). Units within this area are destroyed with a likelihood determined by the hit probability parameter. In contrast, a rocket deterministically destroys a unit when it reaches an aircraft's position, with an error margin of approximately the length of the aircraft, i.e., \qty{10}{\metre}. The position of an aircraft in the map is determined from the aircraft's center point.

\begin{table}[htb]
\tbl{Aircraft Control Parameters.}
{\begin{tabular}{m{2.5cm}m{1.5cm}m{1.2cm}m{1.2cm}m{1.2cm}}
\hline
\textbf{Parameter} & \textbf{Symbol} & \textbf{Unit} & \textbf{AC1} & \textbf{AC2} \\ \hline 
Angular Velocity & $\omega_{\mathrm{AC}}$ & [\unit{\degree/s}] & $[\num{0},\num{5}]$ & $[\num{0},\num{3.5}]$ \\
Speed & $v_{\mathrm{AC}}$ & [\unit{\knot}] & $[\num{100},\num{900}]$ & $[\num{100},\num{600}]$ \\
WEZ & $\omega_{\mathrm{WEZ,AC}}$ & [\unit{\degree}] & $[\num{0},\num{10}]$ & $[\num{0},\num{7}]$ \\
Range & $d_{a,\mathrm{AC}}$ & [\unit{\kilo\metre}] & $[\num{0},\num{2}]$ & $[\num{0},\num{4.5}]$ \\
Hit Probability & $p_{\mathrm{AC}}$ & $[\%]$ & $\num{0.70}$ & $\num{0.85}$ \\ \hline
\end{tabular}}
\label{table:aircraft_dynamics}
\end{table}

\subsection{Multi-Agent Reinforcement Learning}\label{sec:MARL_def}
\begin{figure}[htb]
\centerline{\includegraphics[scale=0.17]{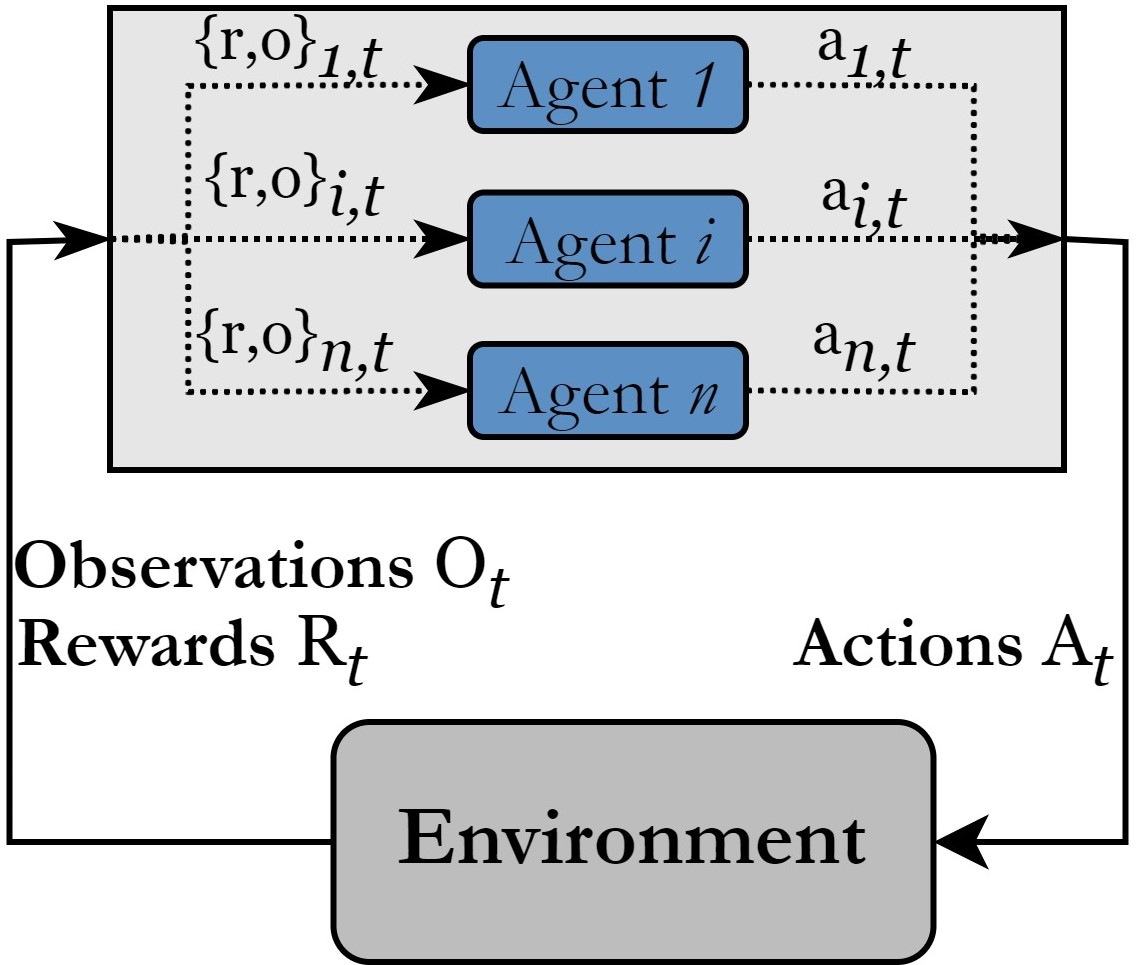}}
\caption{MARL interaction cycle. $R_{i,t}$ denotes rewards, $o_{i,t}$ observations and $a_{i,t}$ are actions.}
\label{fig:MARL_interaction}
\end{figure}

RL is a computational approach used to solve sequential decision-making problems. In MARL, there are multiple agents interacting in a shared environment either in a cooperative or competing fashion, or both. Fig.~\ref{fig:MARL_interaction} illustrates the interaction cycle. As a mathematical framework for modeling the interactions of $n$ agents, we use a \emph{Partially-Observable Markov Game} (POMG)~\cite{pomg}, which is defined by the tuple:
$$(\mathcal{N}, \mathcal{S}, \{\mathcal{O}^i\}_{i=1}^n, \{\mathcal{A}^i\}_{i=1}^n, P,\{R^i\}_{i=1}^n, \gamma, \rho_0).$$
POMGs generalize Markov-Decision Processes~\cite{mdp_bellman} to multiple agents as follows:  

\begin{itemize}
    \item $\mathcal{N}=\{1, \dots, n\}$ is the set of agents, with $i\in\mathcal{N}$ being an agent index.
    \item $\mathcal{S}$ is the state-space representing possible configurations of the environment.
    \item $\mathcal{O}^i \subset \mathcal{S}$ is the observation-space for agent $i$.
    \item $\mathcal{A}^i$ is the set of actions for player $i$, with $\boldsymbol{\mathcal{A}} = \prod_i \mathcal{A}^i$ being the joint action-space.
    \item $P: \mathcal{S} \times \boldsymbol{\mathcal{A}} \to \Delta(\mathcal{S})$ is the state transition probability kernel.
    \item $R^i: \mathcal{S} \times \boldsymbol{\mathcal{A}} \times \mathcal{S} \to \mathbb{R}$ is the reward function of agent $i$.
    \item $\gamma \in [0,1)$ is the discount factor, used to discount future rewards.
    \item $\rho_0 \in \Delta \mathcal{S}$ is the initial state distribution.
\end{itemize}

Starting from an initial state $s_0 \sim \rho$, each agent $i$ gets a local observation $o_t^i \in \mathcal{O}_i$ and selects an action $a_t^i \in \mathcal{A}^i$ according to its individual policy $\pi^i(a \mid o_t^i)$. As other agents simultaneously chose their actions, a joint action $\boldsymbol{a}_t = (a_t^1, \dots, a_t^n) \in \boldsymbol{\mathcal{A}}$ is formed, which is drawn from the joint policy $\boldsymbol{\pi}( \boldsymbol{a}\mid o_t) = \prod_{i=1}^n \pi^i(a\mid o_t^i)$. The POMG then transitions to a new state $s_{t+1}$ according to $P(s_{t+1} \mid s_t, \boldsymbol{a}_t)$, and each agent $i$ receives a reward $R^i(s_t, \boldsymbol{a}_t, s_{t+1})$. The objective in MARL is to find policies $\pi^i$ that maximize the expected return while considering the policies of other agents, as follows:
\begin{equation}
    \pi^i = \arg\max_{\pi} \mathbb{E}\left[\sum_{t=0}^\infty \gamma^t R^i(s_t,\boldsymbol{a}_t,s_{t+1}) \mid \pi^i, \boldsymbol{\pi^{-i}} \right],
\label{eq:joint_policy_optimization}
\end{equation}
where $\boldsymbol{\pi^{-i}}$ represents the policies of all agents except $i$. The challenge in finding the optimal joint policy $\boldsymbol{\pi^*}$ lies in the fact that each agent's reward and thus its optimal policy can depend on the actions of other agents. This interdependence creates a situation where the environment becomes non-stationary from the perspective of any single agent as the policies of other agents evolve.

To mitigate the non-stationary environment problem, we adopt a \emph{Centralized Training with Decentralized Execution}~(CTDE)~\cite{CTDE} scheme for training agents. CTDE is recognized as the state-of-the-art framework for training in multi-agent settings~\cite{Gronauer2021}. Its popularity is attributed to its capability to address non-stationarity, enhance coordination, and preserve each agent’s ability to act independently during execution. In CTDE, agents are trained in a centralized manner where global information is accessible, enabling the learning of coordinated strategies that could be challenging under decentralized training. During execution, each agent operates based on its local observations, ensuring scalability and adaptability. Our modeled POMG with CTDE scheme is used to train control policies. In alignment with the defense domain, which dictate that different maneuvering tactics be employed based on the evolving tactical situation, we define two specific control policies: a \emph{fight} policy, denoted as $\pi_f$, and an \emph{escape} policy, denoted as $\pi_e$. Further on, there is a distinct policy for each aircraft type. Overall we have four control policies: 
\begin{equation}
    \pi \in \{ \pi_{f,\mathrm{AC1}}, \pi_{f,\mathrm{AC2}}, \pi_{e,\mathrm{AC1}}, \pi_{e,\mathrm{AC2}} \}.
    \label{eq:low_level_policies}
\end{equation}

Agents of the same type use the same shared policies. Thus all AC1 use $\pi_{f,\mathrm{AC1}}$ and $\pi_{e,\mathrm{AC1}}$, irrespective on the number of agents, and similarly for AC2. In this way, policies are trained with experiences of all agents of the same type, enabling faster training convergence by eliminating the need to learn distinct policies for each agent. It further enhances coherent behavior across agents within the same team. Coherent behavior in this context is defined as agents $i$ and $j$ of the same type possessing equivalent combat capabilities due to their identical knowledge and maneuvering skills. Additionally, this uniformity simplifies strategic planning for the commander, who can issue commands without considering differences between these agents.

\subsection{Hierarchical Reinforcement Learning}\label{sec:HRL_def}
HRL enhances efficiency in learning and decision-making by employing temporal abstraction. This approach decomposes the overall task into a nested hierarchy of sub-tasks. Abstract commands are issued from higher hierarchy levels to apply a control policy over a limited time span. Symmetries within a particular lower hierarchy level can be exploited by applying the same command to different sub-tasks, such as controlling similar types of airplanes. This enhances scalability by reducing the dimensions of state and action spaces and improves generalization by enabling the generation of new skills through the combination of sub-tasks~\cite{hieR_adv}. Just as in the field of armed forces where strategic directives from higher command levels dictate the operational tactics to lower levels, HRL allows for an organized and systematic approach to handling complex tasks by breaking them down into manageable subtasks, each controlled by a different layer of the learning model. 

Formally, our hierarchical system is modeled as a \emph{Partially Observable Semi-Markov Decision Process} (POSMDP). Our POSMDP includes \emph{options}, which expand the standard concept of actions to include temporally extended courses of action. These options can be thought of as macros (commands), that consist of multiple primitive (control) actions. Options last for varying lengths of time, which is characteristic of Semi-Markov processes, where transitions between states do not necessarily occur at regular time intervals. The POSMDP is defined by the tuple $$(\mathcal{S}, \mathcal{O}, \mathcal{A}, \mathcal{T}, P, R, T_l, \gamma, \rho_0),$$ where:

\begin{itemize}
    \item $\mathcal{S}$ is the set of states of the environment.
    \item $\mathcal{O} \subset \mathcal{S}$ is the set of observations.
    \item $\mathcal{A}$ is the set of actions available at each state.
    \item $\mathcal{T}$ is the set of options, where each option $\tau \in \mathcal{T}$ is defined by a triplet $(I_\tau, \pi_\tau, \beta_\tau)$:
    \begin{itemize}
        \item $I_\tau \subseteq \mathcal{S}$ is the initiation set, specifying the states from which the option can be initiated.
        \item $\pi_\tau(a \mid o)$ is the policy associated with the option, responsible for action selection $a \in \mathcal{A}$, i.e., $\pi_\tau$ is one of the low-level control policies (Eq.~\eqref{eq:low_level_policies}).
        \item $\beta_\tau(s)$ is the termination condition, specifying the probability of the option terminating at each state.
    \end{itemize}
    \item $P: \mathcal{S} \times \mathcal{T} \rightarrow \Delta(\mathcal{S})$ is the state transition kernel that defines the probability of landing in state $s'$ from state $s$ after the execution of option $\tau$.
    \item $R: \mathcal{S} \times \mathcal{T} \times \mathcal{S} \to \mathbb{R}$ is the reward function, providing the immediate reward after transitioning from state $s$ to $s'$ when executing $\tau$.
    \item $T_l:\mathcal{S} \times \mathcal{A} \rightarrow \mathbb{R}$ specifies the execution time for an option $\tau$.
    \item $\gamma \in [0,1)$ is the discount factor, used to discount future rewards.
    \item $\rho_0 \in \Delta \mathcal{S}$ is the initial state distribution.
\end{itemize}

\begin{figure}[htbp]
\centerline{\includegraphics[scale=0.09]{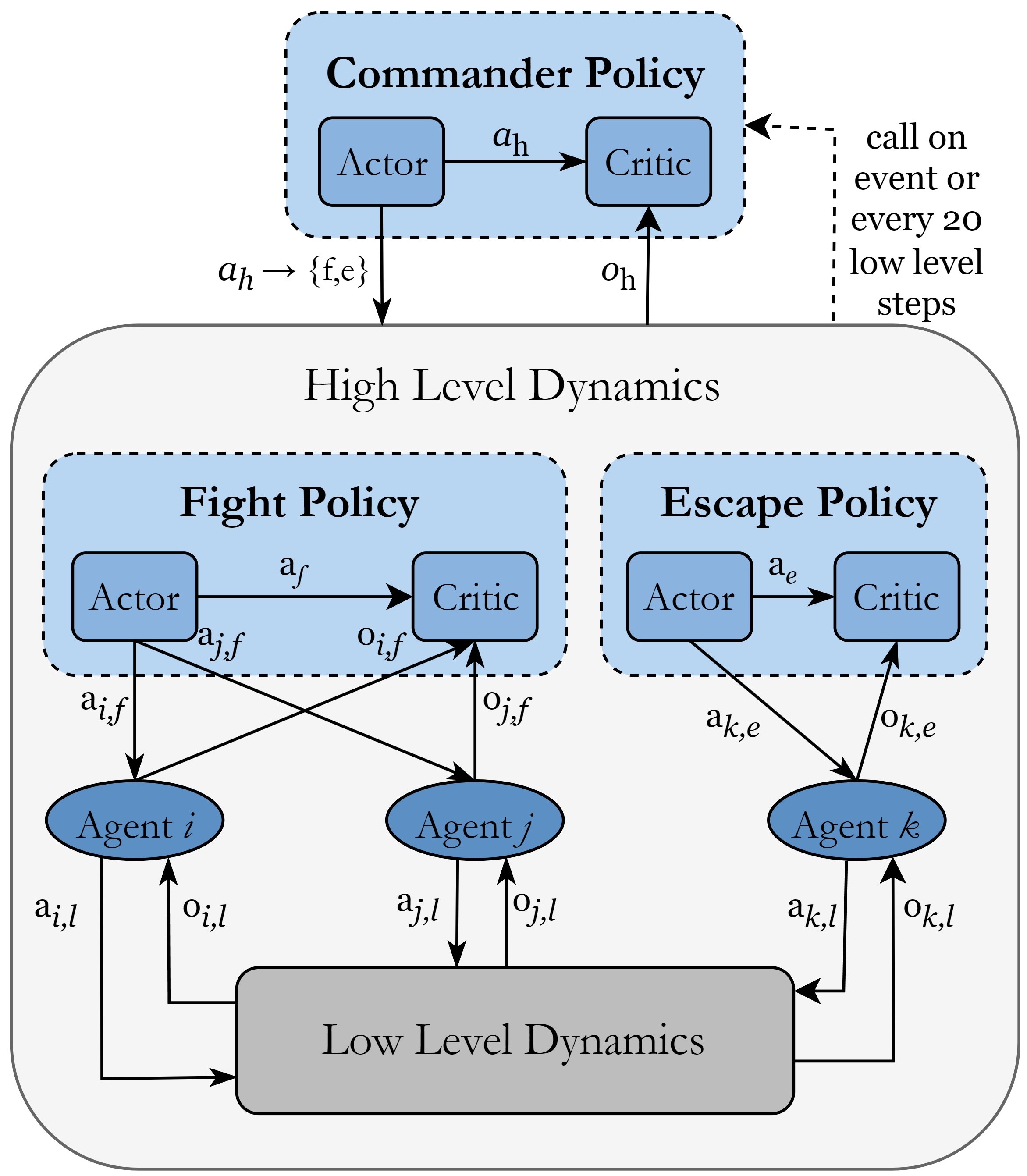}}
\caption{Hierarchy of policies. The commander operates at the high-level dynamics for strategic planning by observing more information of the current situation, while fight and escape policy are deployed to control the aircraft at low-level dynamics. Each policy is equipped with an Actor for execution and a Critic for Learning.}
\label{fig:hierarchy_policy}
\end{figure}

The objective is to maximize the expected return, similarly to Eq.~\eqref{eq:joint_policy_optimization}. We again use a shared CTDE approach to train a single high-level commander policy $\pi_c$ to be used for all agents and aircraft types. Therefore, the commander does not differentiate between AC1 and AC2, since each agent is trained to handle any combat scenario. This setting reflects the concept of policy symmetries, where the same strategy can effectively be applied to any agent. While exploiting policy symmetries can render the commander's strategic behavior homogeneous, the actual operations and capabilities of individual agents of different types maintain their heterogeneity. Fig.~\ref{fig:hierarchy_policy} illustrates the relations between high-level commander policy and low-level control policies (as defined in Sec.~\ref{sec:MARL_def}). The commander policy operates at the high-level dynamics, where it is provided with broader information of the scene for strategic planning. The low-level control policies are directly responsible for controlling the aircraft to ensure successful engagement. Depending on the chosen option $\tau$ by the commander, one of the low-level policies $\pi_f$ or $\pi_e$ is activated per agent and executed during $T_l$ time-steps or until the termination condition $\beta_\tau$ is met. When an option terminates, the commander reassess the situation and determines new tactics. 

\section{Method} \label{sec:method}
This section presents the training methodology of our hierarchical MARL approach, including a structural overview, as well as detailed explanations of each policy, architecture, and algorithm.

\subsection{Structural Overview}\label{sec:overall_structure}
\begin{figure}[htb]
\centering
\subfloat[Hierarchical MARL training loop.]{%
\resizebox*{6.5cm}{!}{\includegraphics{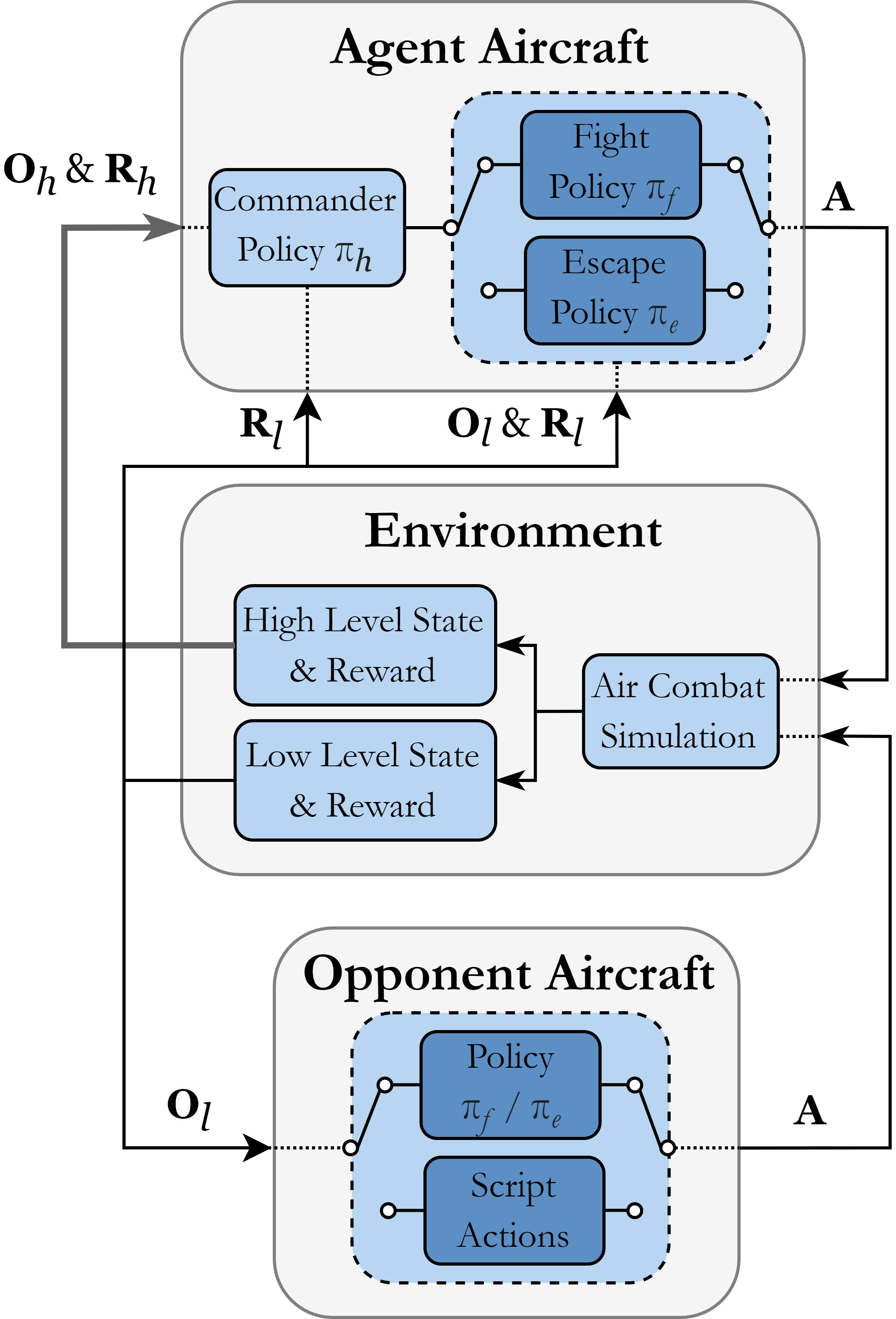}
\label{fig:hhmarl_trainig_loop}}}
\hspace{10pt}
\subfloat[Neural network architecture.]{%
\resizebox*{6.5cm}{!}{\includegraphics{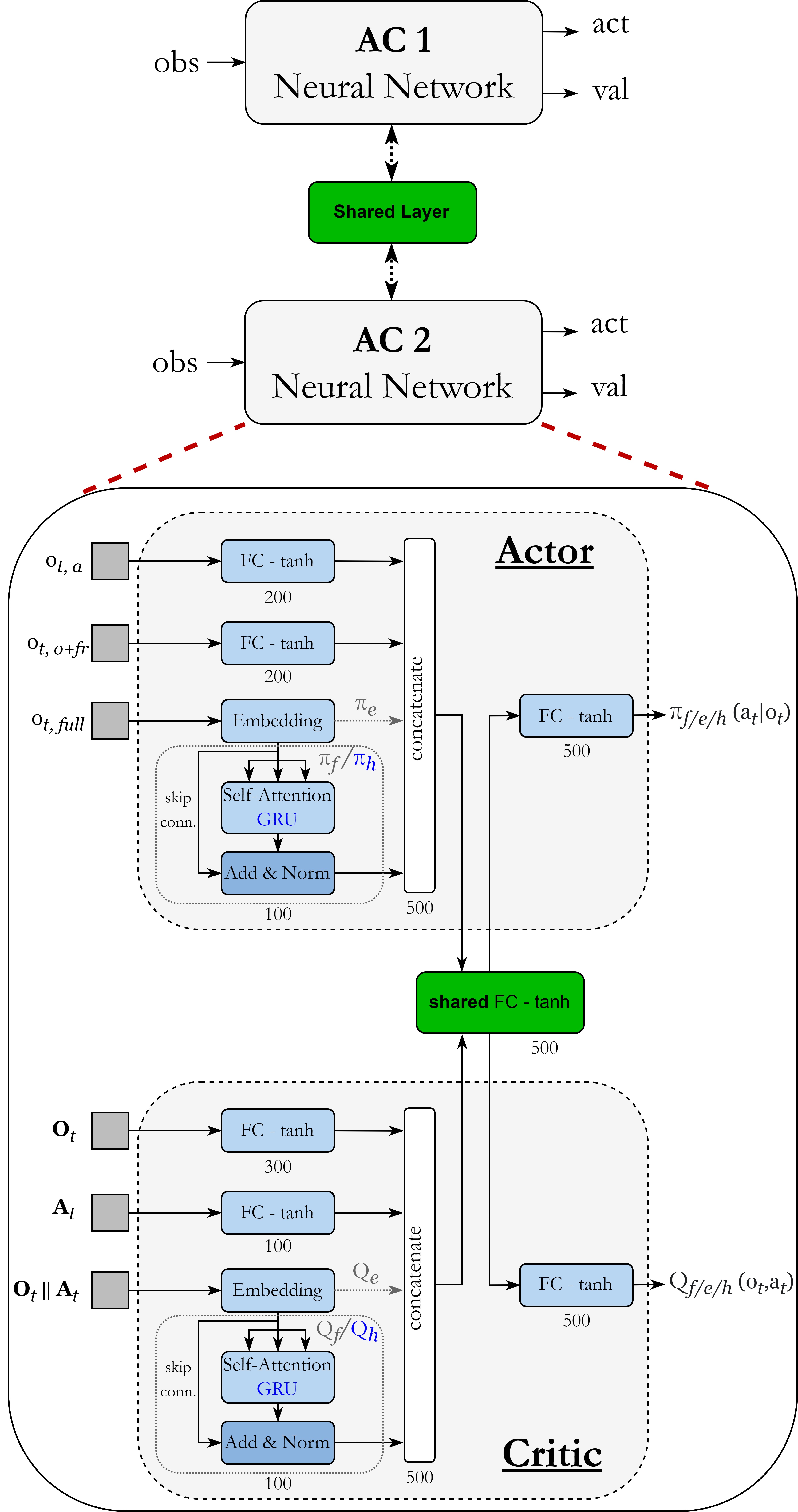}
\label{fig:network_architecture}}}
\caption{The main components representing our approach: (a) Within the environment, we differentiate between high- and low-level information, depending on which policy is being trained. Agents are equipped with either $\pi_f$ or $\pi_e$, and the commander decides which to activate. Opponent aircraft are not controlled by a commander. (b) Each aircraft type operates on its own network instance. The green layer is shared between both instances, as well as between their actors and critics. } 
\label{hhmarl_setup}
\end{figure}

The training loop of our hierarchical MARL algorithm is divided into two main stages, as illustrated in Fig.~\ref{fig:hhmarl_trainig_loop}. Initially, we train the low-level policies, $\pi_f$ and $\pi_e$, utilizing observations $O_l$ and rewards $R_l$. Within the dashed boxes, one type of the control policy (fight or escape) is activated based on the current level of training, where a scripted behavior only applies to opponents. In the subsequent stage, these low-level policies are fixed, meaning no further learning occurs. They serve as options for the high-level commander, $\pi_c$, as defined in Sec.~\ref{sec:HRL_def}. This commander policy is trained using observations $O_h$ and a combined reward signal $R_h + R_l$. The commander controls the activation switches within the dashed areas to select the appropriate policy for the agents. For opponents, the frequency of selecting $\pi_f$ and $\pi_e$ is predetermined manually.

The training of the low-level policy $\pi_f$ is structured into five levels according to a curriculum learning strategy. The complexity of each level is incrementally increased by combating against more competitive opponents. Table~\ref{table:curR_levels} defines the opponent behaviors at each stage. When training is completed at a level, we transfer the policy to the next level and continue training. The escape policy $\pi_e$ is trained directly on L3 and afterwards against $\pi_f$ of L5. The design choices are motivated as follows: 

\begin{itemize}
    \item In L1, the agent is tasked with the fundamental objective of approaching opponents and destroy them. To simplify the initial learning phase, opponents are static, i.e., hovering over a fixed position.
    \item Upon completion of L1, the training progresses to L2, where the opponents begin to move randomly, challenging the agents to learn pursuit and tracking skills.
    \item Starting from L3, the core combat scenarios start, requiring the agents to master precise maneuvering techniques, with the difficulty increasing progressively from L3 to L5. Scripted opponents are programmed to engage the closest agent and to randomly escape. Further details are given in Sec.~\ref{sec:rule-based_opps}.
\end{itemize}

\begin{table}[htb]
\tbl{Curriculum Learning Levels}
{\begin{tabular}{m{2.3cm}m{2.3cm}m{2.3cm}m{2.3cm}m{2.3cm}}
\hline
\textbf{Level 1 (L1)} & \textbf{Level 2 (L2)} & \textbf{Level 3 (L3)} & \textbf{Level 4 (L4)} & \textbf{Level 5 (L5)} \\ \hline 
Static hovering, fixed position & Random maneuvers and firing & Rule-based combat strategies & previously learned L3 Policy & on of previously learned Policies L1-L4 \\ \hline
\end{tabular}}
\label{table:curR_levels}
\end{table}

Our neural network is based on Actor-Critic~\cite{ac_net} (Fig.~\ref{fig:network_architecture}). Low-level AC1 and AC2 agents possess distinct neural network instances (with different input and output dimensions). This means, aircraft of the same \emph{type} utilize a shared instance of the neural network, i.e., a shared policy. Additionally, both instances incorporate a shared layer (indicated by the green box), which is utilized by both the actor and the critic components within each network. This parameter sharing mechanism can enhance coordination and collaborative decision-making among agents~\cite{param_sharing}. The architecture modifications are marked for the three policy types. The policy $\pi_f$ incorporates a \emph{Self-Attention} (SA) module~\cite{attention}, while $\pi_c$ employs a \emph{Gated-Recurrent-Unit} (GRU) module~\cite{gru}. $\pi_e$, does not utilize either module. The embedding layer is configured with a linear layer comprising $100$ neurons and utilizes a $tanh$ activation function. The high-level commander policy $\pi_c$ has only one network instance, irrespective of the aircraft types. As we employ the CTDE approach, the critic receives inputs consisting of the observations and actions of all interacting agents (global information). In addition to parameter sharing, the presence of a fully observable critic enhances coordination among heterogeneous agents~\cite{Gronauer2021}. We update our network parameters using the Actor-Critic approach of Proximal Policy Optimization (PPO)~\cite{ppo} (see Algorithm~\ref{ppo_algo}). The PPO objective seeks to maximize the expected return over trajectories by optimizing a clipped surrogate objective function to ensure stable policy updates. We have consistently kept the PPO algorithm to ensure simplicity in implementation and capitalize on its widespread recognition within the RL community. PPO is recognized for its flexibility in various environments, effectiveness in achieving superior decision-making and robust performance even under challenging conditions and multi-agent games~\cite{mappo}.

\begin{algorithm}[htb]
\caption{PPO training procedure for $\pi_f$, $\pi_e$ and $\pi_c$}
\algsetup{linenosize=\small}
\small
\begin{algorithmic}[1]
\STATE Set number of episodes $N$, time horizon $T$, batch size $b$ and levels $L$
\STATE Initialize buffer $D$ $\leftarrow$ $\{\}$, policy parameters $\theta$, value function parameters $\phi$
\FOR{level $l=1$ to $L$}
\FOR{episode $n=0$ to $N$}
\STATE Initialize state $S_0$
\FOR{$t=0$ to $T$}
\STATE Get agent actions $A_{t,ag}$ by current policy $\pi_{\theta}$
\STATE Get opp action $A_{t,o}$: script if $l\leq3$ else $\pi_{\theta_{l-1}}$
\STATE Execute $(A_{t,ag}, A_{t,o})$, obtain $R_t$ and $S_{t+1}$
\STATE Store $D$ $\leftarrow$ $(S_t, A_{t,ag}, A_{t,o}, R_t, S_{t+1})$
\ENDFOR 
\IF{$|D| \geq b$}
\STATE compute advantage estimation $A^{\pi_{\theta}}$ 
\FOR{update iteration $k=1$ to $K$}
\STATE update policy parameters \newline
$\theta_{k+1} = \underset{\theta}{\arg\max} \mathop{\mathbb{E}}_{\tau \sim D}[\sum_{t=0}^{T}[\min( \frac{\pi_{\theta}(a_t|s_t)}{\pi_{\theta_k}(a_t|s_t)}A^{\pi_{\theta_k}},$ $clip(\frac{\pi_{\theta}(a_t|s_t)}{\pi_{\theta_k}(a_t|s_t)}, 1-\epsilon, 1+\epsilon)A^{\pi_{\theta_k}})]]$
\STATE update value function parameters \newline
$\phi_{k+1} = \underset{\phi}{\arg\min} \mathop{\mathbb{E}}_{\tau \sim D} [\sum_{t=0}^{T} (V_{\phi}(s_t)-\hat{R}_t)^2]$
\ENDFOR
\STATE empty buffer $D$ $\leftarrow$ $\{\}$
\STATE update parameters: $\theta$ $\leftarrow$ $\theta_{k+1}$, $\phi$ $\leftarrow$ $\phi_{k+1}$
\ENDIF
\ENDFOR 
\ENDFOR 
\end{algorithmic}\label{ppo_algo}
\end{algorithm}

\subsection{Air Combat Metrics}\label{sec:metrics}
\begin{figure}[htb]
\centering
\subfloat[]{%
\resizebox*{2cm}{!}{\includegraphics{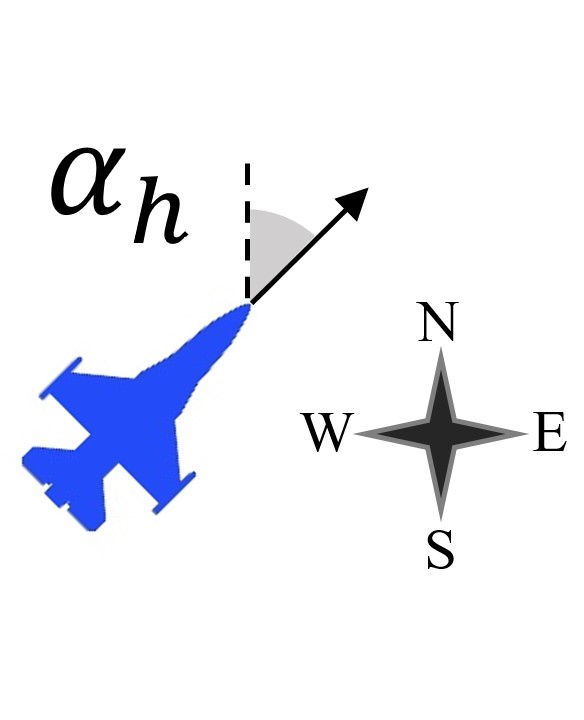}}\label{fig:heading_north}}\hspace{5pt}
\subfloat[]{%
\resizebox*{2cm}{!}{\includegraphics{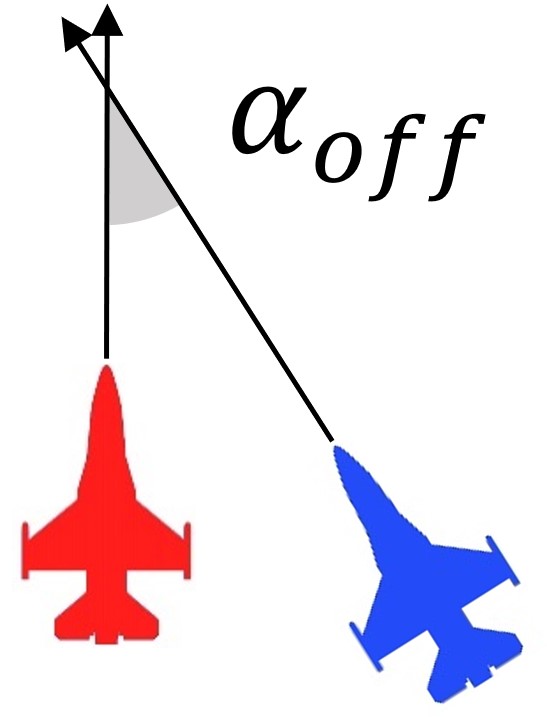}}}\hspace{5pt}
\subfloat[]{%
\resizebox*{2cm}{!}{\includegraphics{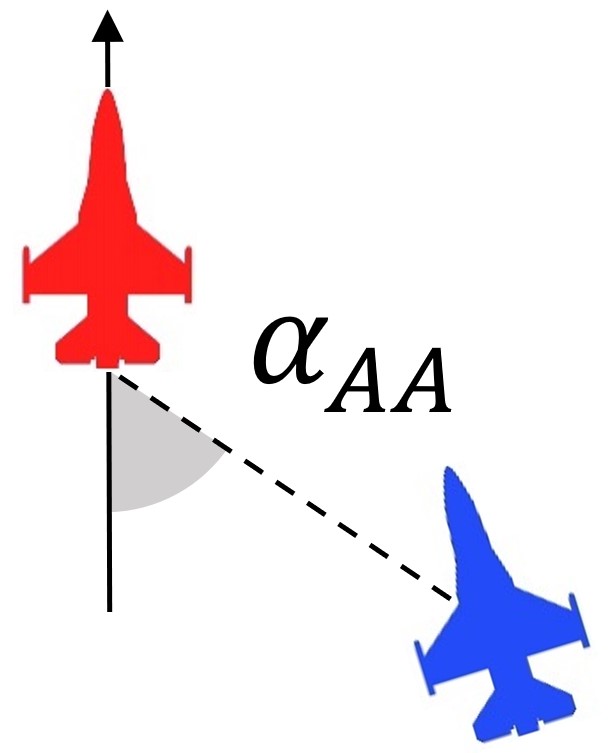}}}\hspace{5pt}
\subfloat[]{%
\resizebox*{2cm}{!}{\includegraphics{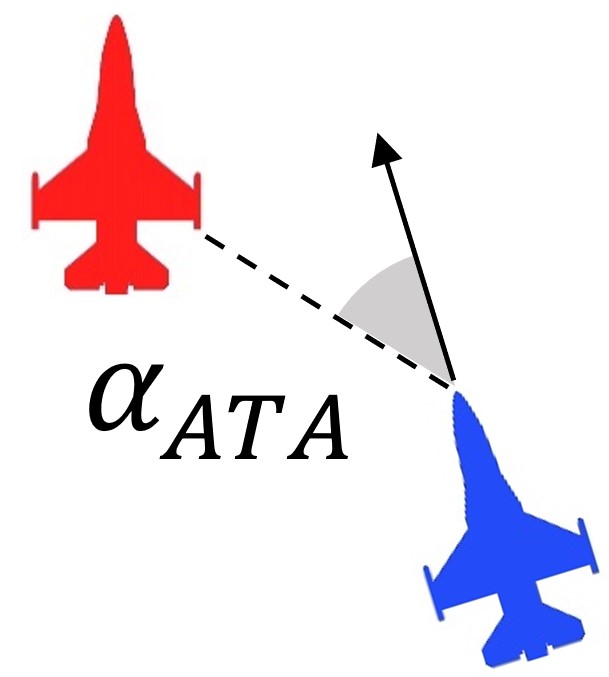}}}\hspace{5pt}
\subfloat[]{%
\resizebox*{2cm}{!}{\includegraphics{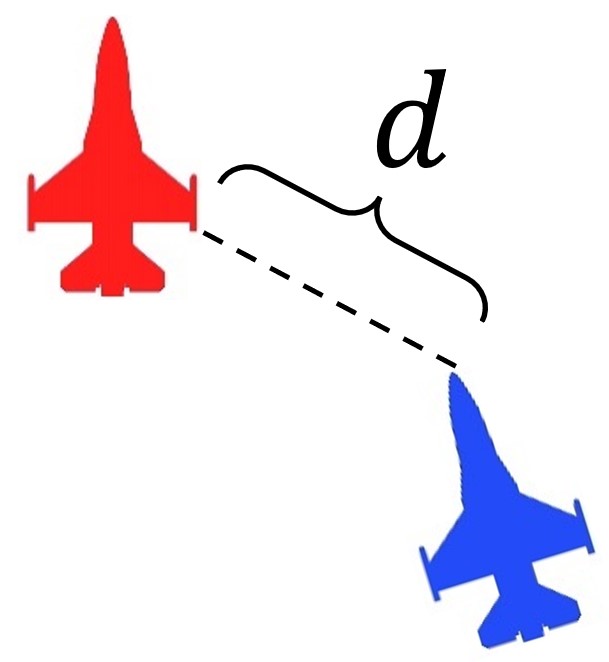}}}
\caption{Aircraft metrics: (a) heading, (b) heading off, (c) aspect angle, (d) antenna train angle, (e) distance.} 
\label{fig:ac_metrics}
\end{figure}

Before we delve into the definition of each policy, we shortly describe the metrics involved. The observation values defining the orientation to opponents are illustrated in Fig.~\ref{fig:ac_metrics}. The actual heading angle $\alpha_h$ is defined w.r.t. north direction. The difference in heading angles between an agent and opponent is captured by $\alpha_{\mathrm{off}}$. The aspect angle $\alpha_{\mathrm{AA}}$ is defined as the angle from the opponent's tail to the position of the agent aircraft, whereas the antenna train angle $\alpha_{\mathrm{ATA}}$ is the difference of the agent's current heading direction to the opponent's position. The parameter $d$ measures the actual distance between the center points of two aircraft. Further observations include map position ($x,y$), current speed ($s$), remaining cannon ammunition ($c_1$) and remaining rockets ($c_2$). Indicator ($w$) defines if the next rocket is ready to be fired and ($s_r$) indicates if the aircraft is currently shooting. Subscript $a$ indicates agent, $o$ opponent and $fr$ friendly aircraft (i.e., from the same team). A subscript in a value, e.g., $\alpha_{\mathrm{off},o}$, defines the angle-off w.r.t. to the opponent. All observation values are normalized to fall within the range $[\num{0},\num{1}]$ and actions of all policies are discrete.

\subsection{Fight Policy}\label{sec:fight_policy}
The objective of the fight policy $\pi_f$ is to optimally control the aircraft for successful attacking maneuvers. We designed the observation of $\pi_f$ to observe its closest opponent and closest friendly aircraft, where $||$ denotes vector concatenation:
\begin{eqnarray*}
o_{t,a} &:=& [x, y, s, \alpha_h, \alpha_{\mathrm{off}, o}, \alpha_{\mathrm{AA}, o}, \alpha_{\mathrm{ATA}, o}, d_o, c_1, \overbrace{c_2, w}^\text{AC1}, s_r]\\
o_{t,o} &:=& [x, y, s, \alpha_h, \alpha_{\mathrm{off}, a}, \alpha_{\mathrm{AA}, a}, \alpha_{\mathrm{ATA}, a}, d_a, s_r]\\
o_{t, fr} &:=& [x, y, s, \alpha_{\mathrm{ATA}, a}, \alpha_{\mathrm{ATA}, fr}, d_a]\\
o_{t, full} &:=& o_{t, a} || o_{t, o} || o_{t, fr} 
\end{eqnarray*}

The control maneuvers (actions) are defined in Table~\ref{table:actions_fight}:

\begin{table}[htb]
\tbl{Control Actions of $\pi_f$.}
{\begin{tabular}{m{6.7cm}m{4.2cm}}
\hline
\textbf{Parameter} & \textbf{Value} \\ \hline
relative heading maneuvers in range [\ang{-90}, \ang{90}] & $h \in \{\num{-6},\ldots,\num{6}\}, \alpha_h = \num{15} \cdot h + \alpha_h$ \\
velocity mapping of $v$ to ranges of AC1 or AC2 & $v \in \{\num{0},\ldots,\num{8}\}$ \\ 
shooting with cannon & $c \in \{\num{0},\num{1}\}$ \\
shooting with rocket & $r \in \{\num{0},\num{1}\}$ \\ \hline
\end{tabular}}
\label{table:actions_fight}
\end{table}

In air combat, facing the opponent's tail is a favorable situation for shooting. We therefore define the reward function based on $\alpha_{\mathrm{ATA},a}$ of the opponent to the agent. We further encourage the combat efficiency by incorporating the remaining ammunition ($c_{\mathrm{rem}} = c_1 + c_2$), yielding the following killing reward:
\begin{equation}
R_{k} = \alpha_{\mathrm{ATA},a} + \frac{c_{\mathrm{max}}-c_{\mathrm{rem}}}{c_{\mathrm{max}}} \in [\num{1},\num{2}]\,.
\label{eq:rew_kill}
\end{equation}

Punishing rewards are given when getting destroyed by an opponent $R_{d}=\num{-2}$ or destroying a friendly aircraft $R_{fr}=\num{-2}$ as well as flying out of environment boundaries $R_{b}=\num{-5}$. We explicitly design the reward values such that $R_d$ and/or $R_{fr}$ have a greater negative impact compared to $R_k$. Conversely, it would be counterproductive if the combination of penalties for failure and the rewards for successful kills would still result to a positive total. Given that out-of-boundary events may occur less frequently, we allocate a higher penalty to these to ensure they significantly influence the overall reward. For example, in scenarios where an agent eliminates two opponents but subsequently hits a boundary, the agent should receive a net negative reward. The total reward for training $\pi_f$ is thus given as: 
\begin{equation}
    R_{\mathrm{fight}} = R_{k} + R_{d} + R_{fr} + R_{b}\,.
\label{eq:rew_fight}
\end{equation}

\subsection{Escape Policy}\label{sec:esc_policy}
The ultimate goal of $\pi_e$ is to remain alive during the combat scenario by carefully steering the aircraft and avoiding dangerous situations. $\pi_e$ senses two closest opponents and its closest friendly aircraft as:
\begin{eqnarray*}
o_{t,a} &:=& [x, y, s, \alpha_h, c_1, \overbrace{c_2}^\text{AC1}] \\
o_{t,o} &:=& [x, y, s, \alpha_h, \alpha_{\mathrm{off},a}, \alpha_{\mathrm{ATA},a}, \alpha_{\mathrm{ATA},o}, d_a] \\
o_{t, fr} &:=& [x, y, s, \alpha_{\mathrm{ATA},a}, \alpha_{\mathrm{ATA},fr}, d_a] \\
o_{t, full} &:=& o_{t, a} || o_{t, o_1} || o_{t, o_2} || o_{t, fr}
\end{eqnarray*}

The actions remain same as for $\pi_f$, i.e., fleeing agents can still fire and kill opponents. However, they are not rewarded for this. Instead, we design the reward function in that way to purely survive the battle and stay within the environment boundaries. The definitions of $R_d$, $R_{fr}$ and $R_b$ remain the same as for training $\pi_f$. Therefore, the total non-positive training reward for $\pi_e$ is given as:
\begin{equation}
    R_{\mathrm{esc}} = R_{d} + R_{fr} + R_{b}\,.
\label{eq:rew_esc}
\end{equation}

In the ablation studies in Sec.~\ref{sec:esc_policy_results}, we elaborate why this punishing reward yields a better performance than positively rewarding the agent for escaping.

\subsection{Commander Policy}\label{sec:commandeR_policy}
The task for the high-level commander policy $\pi_c$ is to provide strategic commands to the low-level policies for optimal combat tactics. Since $\pi_c$ is trained through a shared CTDE method, it gets invoked by every agent separately. The individual observations are based on two closest aircraft of the calling low-level agent.
\begin{eqnarray*}
o_{t,a} &:=& [x, y, s, \alpha_h] \\
o_{t,o} &:=& [x, y, s, \alpha_h, \alpha_{\mathrm{off},a}, \alpha_{\mathrm{AA},a}, \alpha_{\mathrm{AA},o}, \alpha_{\mathrm{ATA},a}, \alpha_{\mathrm{ATA},o}, d_a] \\
o_{t, fr} &:=& [x, y, \alpha_{\mathrm{ATA},a}, \alpha_{\mathrm{ATA},fr}, d_a] \\
o_{t, full} &:=& o_{t, a} || o_{t, o_1} || o_{t, o_2} || o_{t, fr_1} || o_{t, fr_2} 
\end{eqnarray*}

The tactics the commander learns dictate which low-level policy each agent must activate. The action set (or options, as defined in Sec.~\ref{sec:HRL_def}) is $a_c \in \{\num{0}, \num{1}, \num{2}\}$, where \num{0} activates $\pi_e$ and $\pi_f$ otherwise. If $\pi_f$ is activated, the chosen option ($1$ or $2$) determines which of the two observable opponents the agent should attack. The agent then gets the corresponding observation for its low-level policy. In our setup, the commander adapts to only these pre-trained low-level policies. Worth to mention is, that our setup allows to alter this sensing strategy quite easily, i.e., the commander can be adjusted to sense three instead of two close opponents and make a decision accordingly. In our ablation studies, we examine how this impact the overall performance.

To train the commander, we design an action assessment reward function $R_{\mathrm{act}}$ that should encourage the commander to exploit favorable situations. Concretely, a favorable situation for an agent is defined as being in a good position to attack the opponent with a high chance of destroying it. Formally, is defined as follows, where the conditions are evaluated for all agents w.r.t. each opponent:

\begin{equation}
R_{\mathrm{act}}=
\begin{cases}
+\num{0.1} & d_o\!<\!\qty{5}{\kilo\metre}\wedge\alpha_{\mathrm{ATA},a-o}\!<\!\ang{15}\wedge a_c\!=\!i_{o}\!\in\mathcal{A}_o\\
+\num{0.1} & d_o\!<\!\qty{5}{\kilo\metre}\wedge\alpha_{\mathrm{ATA},o-a}\!<\!\ang{15}\wedge\alpha_{\mathrm{ATA},a-o}\!>\!\ang{30}\wedge a_c\!=\!0\\
-0.1 & a_c\!=\!i_{o}\!\notin\mathcal{A}_o\\
0 & \mathrm{otherwise}.
\end{cases}
\label{eq:fav_sit_reward}
\end{equation}

In Eq.~\eqref{eq:fav_sit_reward}, $d_o$ defines the distance and $\alpha_{\mathrm{ATA},a-o}$ the antenna train angle, both from the perspective of an agent to an opponent, whereas $\mathcal{A}_o$ denotes the set of active opponents. $R_{\mathrm{act}}$ defines a favorable situation in the first case, that is characterized as having a short enough distance for firing and approximately facing the opponent. If these conditions are met for an agent to an opponent and the commander has chosen the corresponding opponent index $i_o$, for which the favorable situation applies, the commander gets an incentive reward. The same reasoning holds for the second case, if the commander sets the agent into escaping mode while the opponent is in a favorable situation and the agent faces away from the opponent. Contrary, selecting an opponent index $i_o \notin \mathcal{A}_o$ will punish the commander. To complete the reward function for the commander, we also include the killing reward $R_{k}=\num{1}$ if an agent with its low-level policy killed an opponent and $R_{d}=\num{-1}$ if the agent got destroyed. The out-of-boundary reward $R_{b}=\num{-2}$ is considered as well, but the friendly kill punishment is omitted. The total reward for training $\pi_c$ is defined as:
\begin{equation}
    R_c = R_k + R_d + R_b + R_{\mathrm{act}}\,.
\label{eq:total_commaneR_rew}
\end{equation}

Based on the visualization in Fig.~\ref{fig:hierarchy_policy}, an algorithmic description of the commander training procedure is given in Algorithm~\ref{hieR_algo}. The commander gets invoked dynamically when the low-level horizon $T_l$ is reached or on termination events, characterized as:
\begin{itemize}
    \item any aircraft got destroyed (by shooting or hitting map boundary).
    \item an agent approaches the map boundary ($d<\qty{5}{\kilo\metre}$).
    \item an agent \emph{or} an opponent is in favorable situation (the first case of Eq.~\eqref{eq:fav_sit_reward}).
\end{itemize}

\begin{algorithm}[htb]
\caption{Simulation procedure for training commander policy $\pi_c$}
\algsetup{linenosize=\small}
\small
\begin{algorithmic}[1]
\STATE Set high-level time horizon $T_h$, low-level time horizon $T_l$, max number of episodes $N$, opponent fight policy assignment probability $p_o$
\FOR{episode $n=0$ to $N$}
\STATE Sample a combat scenario
\FOR{$t_h=0$ to $T_h$}
\STATE Get Commander actions for all $n$ agents: $A_{h,t_h} = \{a_{h,1}, \ldots, a_{h,n}\}$
\STATE Agents: activate low-level policies $\pi_{a, f/e} \leftarrow a_{h,i}$ 
\STATE Opponents: policy assignment $\pi_{o, f/e} \leftarrow p_o$ 
\FOR{$t=0$ to $T_l$}
\STATE Execute $\pi_f$ or $\pi_e$ for each agent and opponent
\IF{$t\geq 10$ or event}
\STATE \textbf{break}
\ENDIF
\ENDFOR 
\STATE Get $R_{t,h}$ and $S_{t,h}$
\ENDFOR 
\STATE update $\pi_c$ according to Algorithm~\ref{ppo_algo}
\ENDFOR 
\end{algorithmic}
\label{hieR_algo}
\end{algorithm}

\subsection{Rule-Based Opponents}\label{sec:rule-based_opps}
\begin{figure}[htb]
\centerline{\includegraphics[scale=0.28]{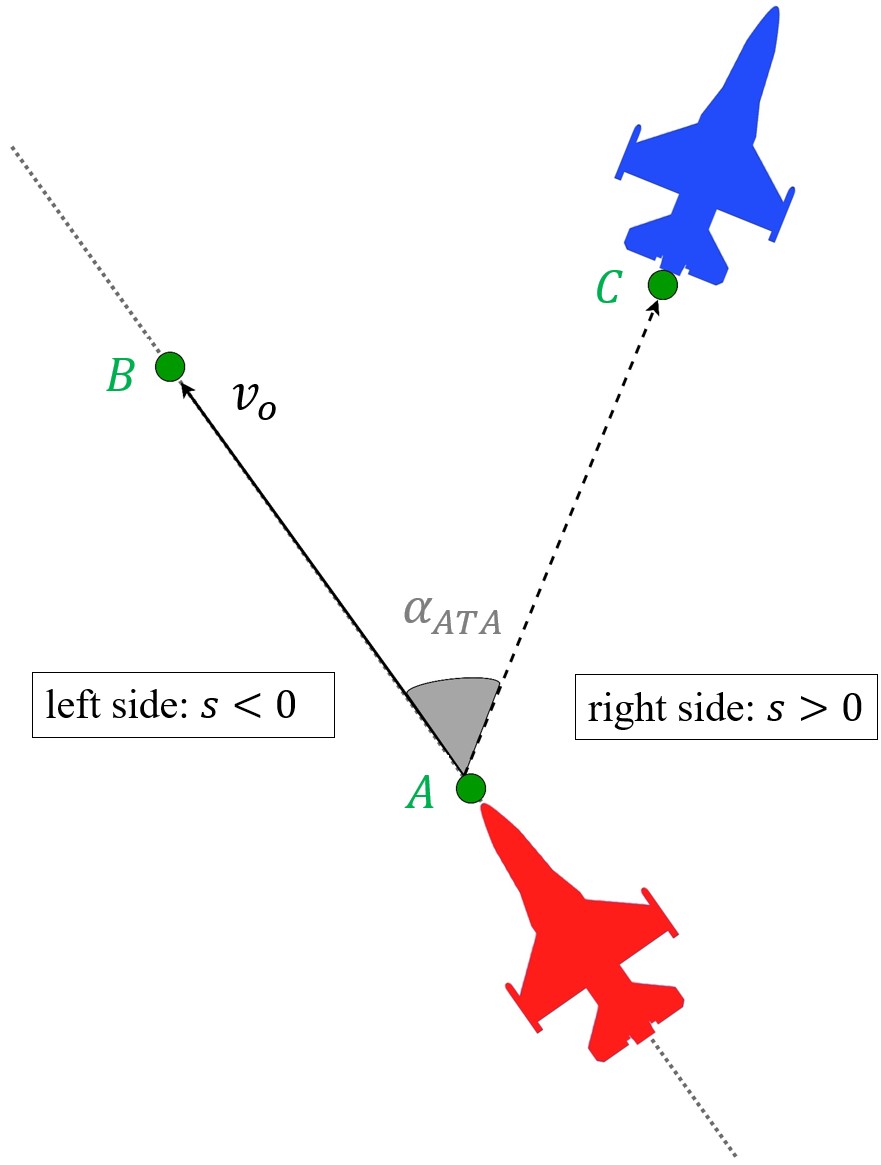}}
\caption{Calculation of turning direction of opponent (red) to face agent (blue). $v_o$ is a unit vector aligned with the heading direction of the opponent. For visualization purposes, the locations of aircraft are moved to the front (point A) and to the tail (point C), but are in fact located at each aircraft's center point. The sign $s$ determines if an agent is located to the right or to the left of an opponent.}
\label{fig:script_turning}
\end{figure}

In the computation of navigation strategies for opponents in L3, our methodology focuses on the closest agent of each opponent, employing the calculation of vector determinant for directional calculations. Given the antenna train angle $\alpha_\mathrm{ATA} \in [\num{0},\num{180}]\unit{\degree}$ as defined in Sec.~\ref{sec:metrics}, which defines the difference of the current heading angle to the target, we ascertain the requisite angular adjustment to align with the target. Direct application of this angular disparity to the opponent's current orientation could facilitate immediate alignment. Specifically, a straightforward calculation of $\alpha_h + \alpha_{\mathrm{ATA}}$ invariably orients the adversaries to the right, as $a_h$ is measured w.r.t. to the northern hemisphere, as depicted in Fig.~\ref{fig:heading_north}. However, this approach is not always optimal, as it might provoke superfluous rotational movements. To mitigate this, we refer to Fig.~\ref{fig:script_turning} and construct a unit vector $v_o$, originating at the opponent's location (red aircraft) and extending in the direction of the current heading angle. Determining whether the target is situated on the left or right side of $v_o$ is crucial for deciding the direction of rotation necessary for rapid alignment. This determination involves the points $A$ and $B$ (origin and terminus of $v_o$, respectively) and point $C$, representing the target's position. Each point is a two-dimensional coordinate featuring both $x$ and $y$ components. The directional decision is influenced by the sign $s$ of the determinant of the vectors $(\overrightarrow{AB},\overrightarrow{AC})$. The sign of the determinant indicates the orientation of the transformed vector space relative to the original. A positive determinant preserves the orientation, while a negative determinant indicates a reversed orientation. This is calculated as follows:
\begin{equation}
s = \text{sign}((B_x - A_x)(C_y - A_y) - (B_y - A_y)(C_x - A_x))\,.
\label{eq:sign_opp}
\end{equation}

This sign $s$ in Eq.~\eqref{eq:sign_opp} determines whether to increment or decrement $\alpha_h$ in $\alpha_h = \alpha_h + s\cdot\alpha_{\mathrm{ATA}}$, or whether to turn left or right to quickly face the agent, respectively. To introduce variability and uncertainty in behavior, the heading calculation is infused with some randomness $r \sim \text{Uniform}(0, 1)$. This yields the final computation formula for determining the heading direction of opponents:
\begin{equation}
\alpha_h = \alpha_h + s\cdot r\cdot \alpha_{\mathrm{ATA}}\,.
\label{eq:script_heading}
\end{equation}

Additional considerations for adversary strategy include velocity modulation based on proximity to the target and decision-making related to shooting actions. Specifically, the likelihood of firing with cannon or rocket increases inversely with $\alpha_{\mathrm{ATA}}$. Furthermore, to avoid entrapment in rotational patterns, opponents occasionally employ random fleeing maneuvers. This framework underpins adversary behavior in L3, leading to the steepest learning effect for agent's combat skills.

\section{Experiments}\label{sec:experiments}
In this section, we begin by describing our simulation setup, which includes the development of our platform and the training configuration. We will then present experimental results and conduct a detailed analysis of the combat performance of each policy. Our code for running simulations is publicly available.\footnote{Official implementation: \href{https://github.com/IDSIA/marl}{https://github.com/IDSIA/marl}.}

\subsection{Simulation Settings}\label{sec:simulation}
\subsubsection{Environment and Libraries}\label{sec:sim_platform_env}
\begin{figure}[htb]
\centering
\subfloat[Commander $\pi_c$.]{%
\resizebox*{4.6cm}{!}{\includegraphics{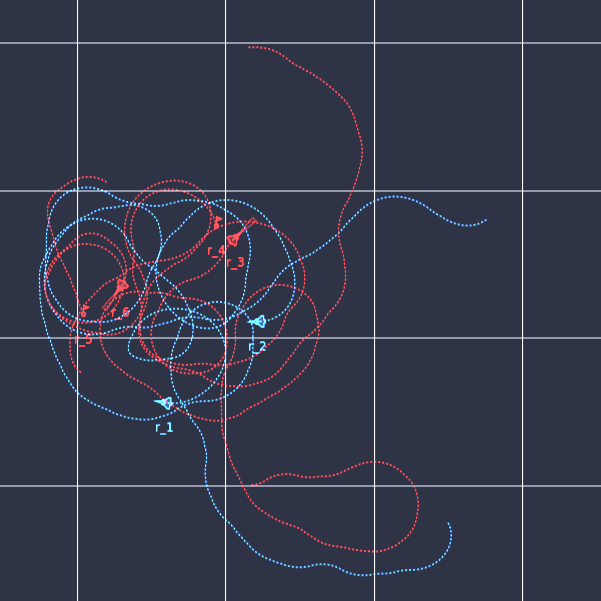}
\label{fig:traj_hier}}}
\hspace{1pt}
\subfloat[Fight $\pi_f$.]{%
\resizebox*{4.6cm}{!}{\includegraphics{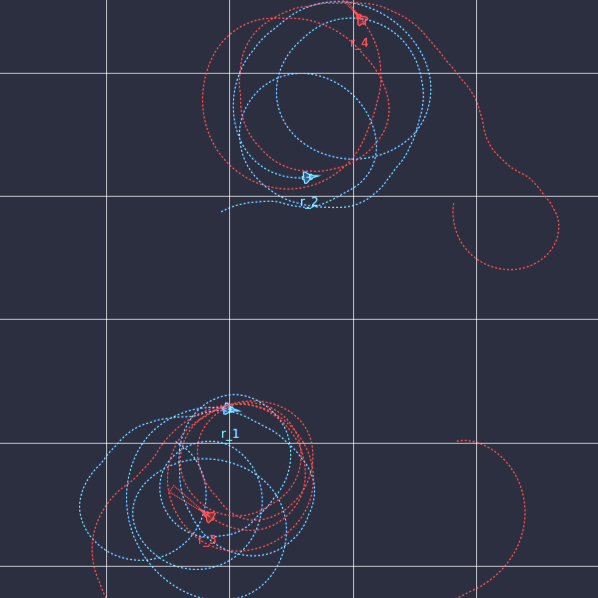}
\label{fig:traj_fight}}}
\hspace{1pt}
\subfloat[Escape $\pi_e$.]{%
\resizebox*{4.6cm}{!}{\includegraphics{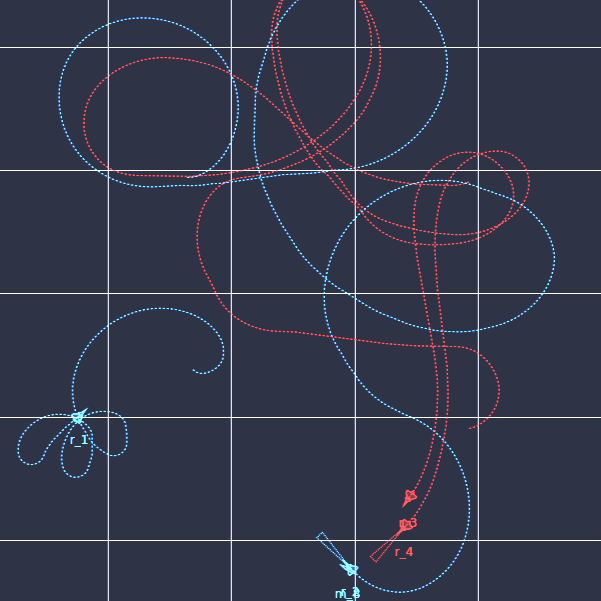}
\label{fig:traj_esc}}}
\caption{Different policy trajectories in the simulation platform.} 
\label{fig:trajectories}
\end{figure}

A notable aspect of our work is the development of a dedicated 2D (Python) simulation platform to have full control and low inertia. The simulator is based on the discrete-event simulation architecture: at each simulation procedure, the next position of the units is computed as if a fixed amount of time has passed (typically $0.1$ seconds). During this time frame, distinct events can occur, such as the firing of a rocket or the explosion of a unit. At the end of each simulation round, a new simulation state is provided, along with the generated events. The dynamics of the units is simplified to a first-order physical simulation: only speed and direction is taken into account. Movements of aircraft are computed with respect of geodesics (WGS84), so even on larger maps the movements are plausible. Our platform is lightweight, fast, and accurately simulates the dynamics of our aircraft for the experiments described in this work. Fig.~\ref{fig:trajectories} shows some rendered scenes of our environment under different combat settings. Trajectories of each aircraft can be visualized and a landmark is set at the position where an aircraft got destroyed. Map size and number of interacting aircraft can be specified, highlighting the diversity and scalability properties of our model. Additionally, it is feasible to model various types of aircraft and integrate them into the training scenario, ensuring that our setup is tailored to reflect a wide range of potential real-world engagements. We refer to time-step $t$ as one simulation round.

\subsubsection{Training and Evaluation Configuration}\label{sec:sim_training_config}

In the base definition of our approach, we use a shared CTDE policy for each aircraft type and for the commander. This configuration enables our method to support simulations with a variable number of agents, as each agent of the same type utilizes the same learned policy for decentralized execution.

A simulation episode ends when either the time horizon is reached or there are no active aircraft of one team. An aircraft is destroyed when getting hit by cannon or rocket or when hitting the map boundary. For each episode, a side of the map (left or right half) is chosen at random for each team, followed by generating random initial positions and headings for each aircraft. To achieve variable team configurations, aircraft types are randomly selected, with at least one of each type included per group to ensure heterogeneity. Map sizes per axis are \qty{30}{\kilo\metre} for low-level and \qty{50}{\kilo\metre} for high-level policy training. Unless otherwise stated, the results are provided as follows: Learning curves showing mean rewards include the training performance of all agents. Since low-level agents can sense one opponent and one friendly aircraft per-time-step, we train them in accordance with their sensing capabilities, i.e., in a 2-vs-2 combat setting. Even though the sensing capabilities of the commander policy $\pi_c$ can be altered to observe two or three opponents at a time, as stated in Sec.~\ref{sec:commandeR_policy}, we configure the training in either sensing strategy to a 3-vs-3 setting. Evaluations are done for \num{1000} episodes and in the same combat setting as in low-level or high-level training, respectively. Low-level training and evaluation comparisons are done for L3, in which the opponents have the most deterministic behavior and therefore enables the most consistent comparison. For inspecting the commander performance, the pre-trained policies $\pi_f$ of L5 and $\pi_e$ are employed. \emph{Win} is when all opponents are destroyed, \emph{Loss} if all agents got destroyed and \emph{Draw} if at least one agent per team remains alive after the episode ends. 

According to Algorithm~\ref{ppo_algo}, we set the PPO parameters as follows, which are kept constant for all training procedures: learning rate for actor and critic is set to $lr=\num{0.0001}$, discount factor $\gamma=\num{0.95}$, clip parameter $\epsilon=\num{0.2}$, Adam as optimizer, batch size of \num{2000} for low-level policies and \num{1000} for high-level policy. We employ the popular libraries \emph{Ray RLlib} and \emph{Pytorch} for training our model.

\subsection{Fight Policy $\pi_f$}\label{sec:fight_policy_results}

The specific aircraft configuration for training $\pi_f$ are set as follows: for every new episode, the aircraft ammunition is set to \num{200} cannon shots (both aircraft types) and \num{5} rockets (only AC1). We make the opponents stronger by giving them ammunition of \num{400} cannons and \num{8} rockets. As the levels increase during the curriculum learning stages, we also increase the time horizon of an episode by $\Delta T=\num{50}$ starting from $T=\num{200}$ on L1. We now thoroughly examine the behavior of $\pi_f$ under different conditions. 

\subsubsection{Reward Inspection}\label{sec:fight_rew_inspect}

We examine the combat behavior when modifying the reward function, which is responsible for defining the overall objective. The first reward function is the base fight reward $R_\mathrm{fight}$ defined in Eq.~\eqref{eq:rew_fight}. The other two functions are: 
\begin{equation}
    R_{\mathrm{FriPun}} = R_{\mathrm{fight}} + R_{fp}, \quad R_{fp} =\num{-2}\,,
\label{eq:FriPun}
\end{equation}
\begin{equation}
    R_{\mathrm{ShFrac}} = R^i_{\mathrm{fight}} + \rho \cdot \sum_{k \in \mathcal{N} \setminus \{i\}} R^k_{\mathrm{fight}}\,.
\label{eq:ShFrac}
\end{equation}

\begin{figure}[htb]
\centering
\subfloat[Training $\pi_f$ with different rewards.]{%
\resizebox*{6cm}{!}{\includegraphics{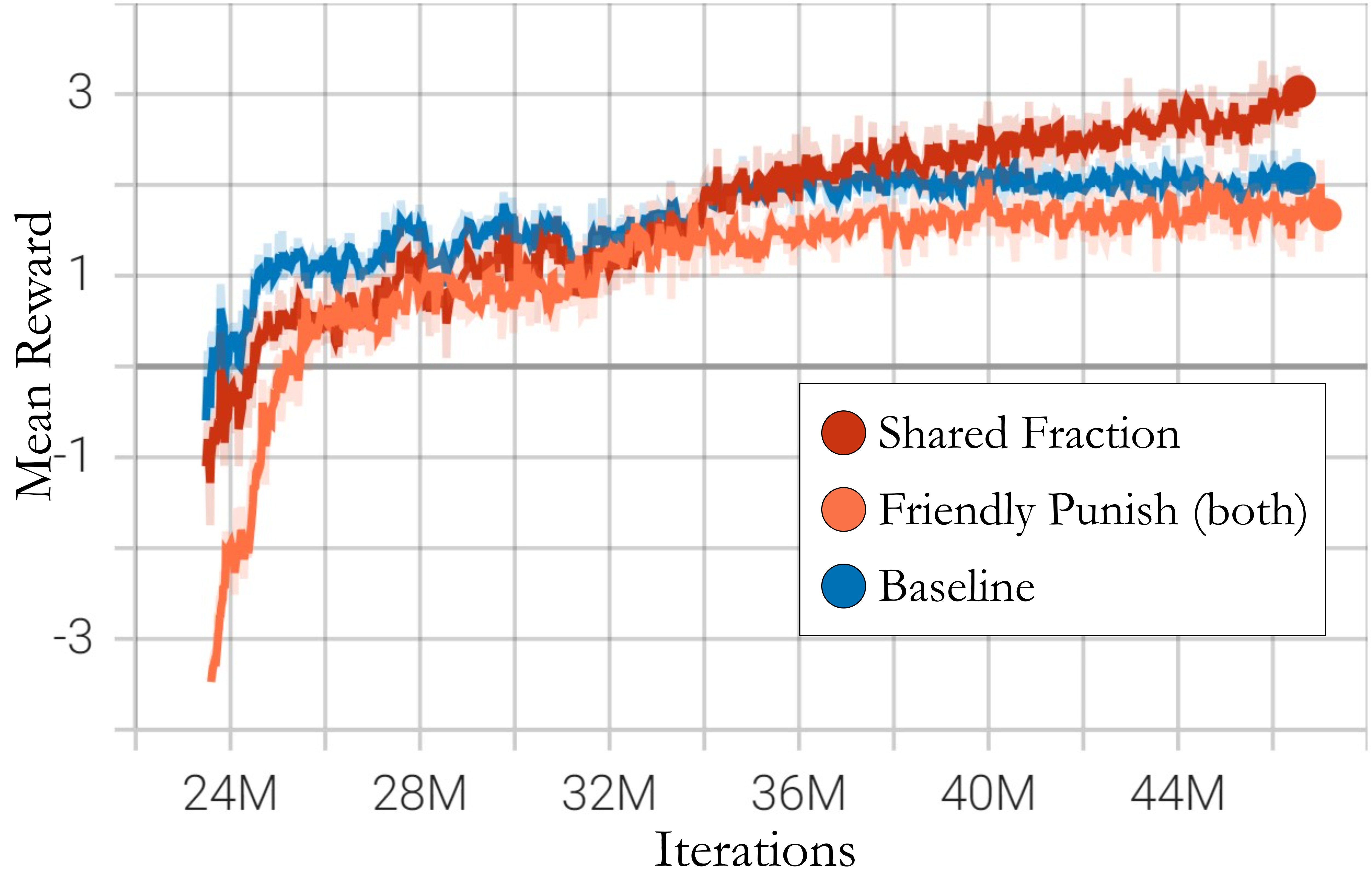}
\label{fig:fight_rews_train}}}
\hspace{5pt}
\subfloat[Evaluation $\pi_f$ with different rewards.]{%
\resizebox*{5.5cm}{!}{
\begin{tikzpicture}
\begin{axis}[
xbar stacked,
legend entries={{\scriptsize Win},{\scriptsize Draw},{\scriptsize Loss}},
bar width=8,
width=4.5cm,
xmajorgrids,
height=3cm,
xmin=0,
xmax=100,
ytick={0,1,2},
yticklabels={{\footnotesize ShFrac},{\footnotesize FriPun}, {\footnotesize Base}},
xtick={0,50,100},
tickwidth=0mm,
legend style={at={(0.5,-0.35)},anchor=north,legend columns=-1},
xticklabels={{\scriptsize $0\%$},{\scriptsize $50\%$},{\scriptsize $100\%$}},
axis on top]
\addplot [color=blue, fill=blue!65!lime] coordinates 
{(56,0) (65,1) (67,2)};
\addplot [color=gray, fill=gray!60!white] coordinates 
{(27,0) (24,1) (19,2)};
\addplot [color=red, fill=red!60!orange] coordinates 
{(17,0) (11,1) (14,2)};
\end{axis}
\end{tikzpicture}
\label{plot:fight_rews_eval}}}
\caption{Comparison on performance under different fight rewards.} 
\label{fig:fight_rews_compare}
\end{figure}
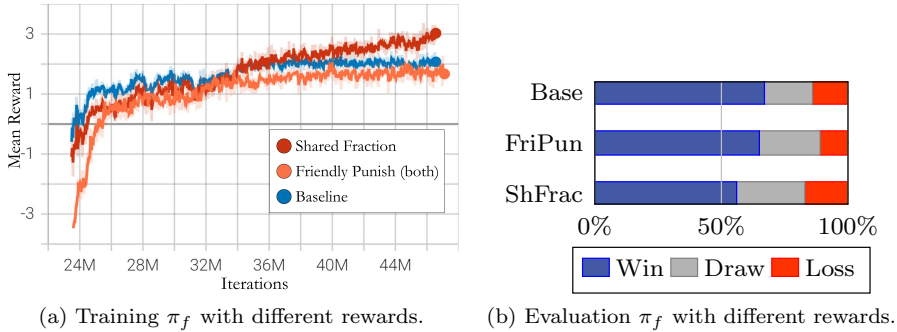

Here, the reward mechanism in Eq.~\eqref{eq:FriPun} is termed Friendly Punishment (abbreviated \emph{FriPun}) because it accounts for scenarios where an agent inadvertently causes damage to a friendly unit. This penalty is already integrated into the base reward $R_{\mathrm{fight}}$, but in this case, it extends to evaluating the impact of such penalties on \emph{both}, the offending ($R_{fr}$) and the victim ($R_{fp}$) agent. This should encourage the agents to exercise greater caution and avoid such incidents. The other reward function defined in Eq.~\eqref{eq:ShFrac} is called Shared Fraction (\emph{ShFrac}) and resembles (partially) the definition of a global reward. In this function, the base fight reward of agent $i$, $R^i_{\mathrm{fight}}$, is augmented by a fraction of the total rewards earned by the remaining friendly agents $\{\num{1}, \num{2}, \ldots, n\} \setminus \{i\}$, controlled by the parameter $\rho$. We designate $\rho = 0.5$ for this ablation study and examine, whether a shared reward might enhance cooperation and performance. The results of training and evaluation across these reward configurations are depicted in Fig.~\ref{fig:fight_rews_compare}. Training data does not decisively indicate which reward mechanism yields superior performance. However, during evaluation, the base individual reward $R_{\mathrm{fight}}$ demonstrates a higher efficacy compared to the alternatives, where the shared reward performed worst.

\begin{figure}[htb]
\centering
\subfloat[Training comparison of AC1 at L3.]{%
\resizebox*{6cm}{!}{\includegraphics{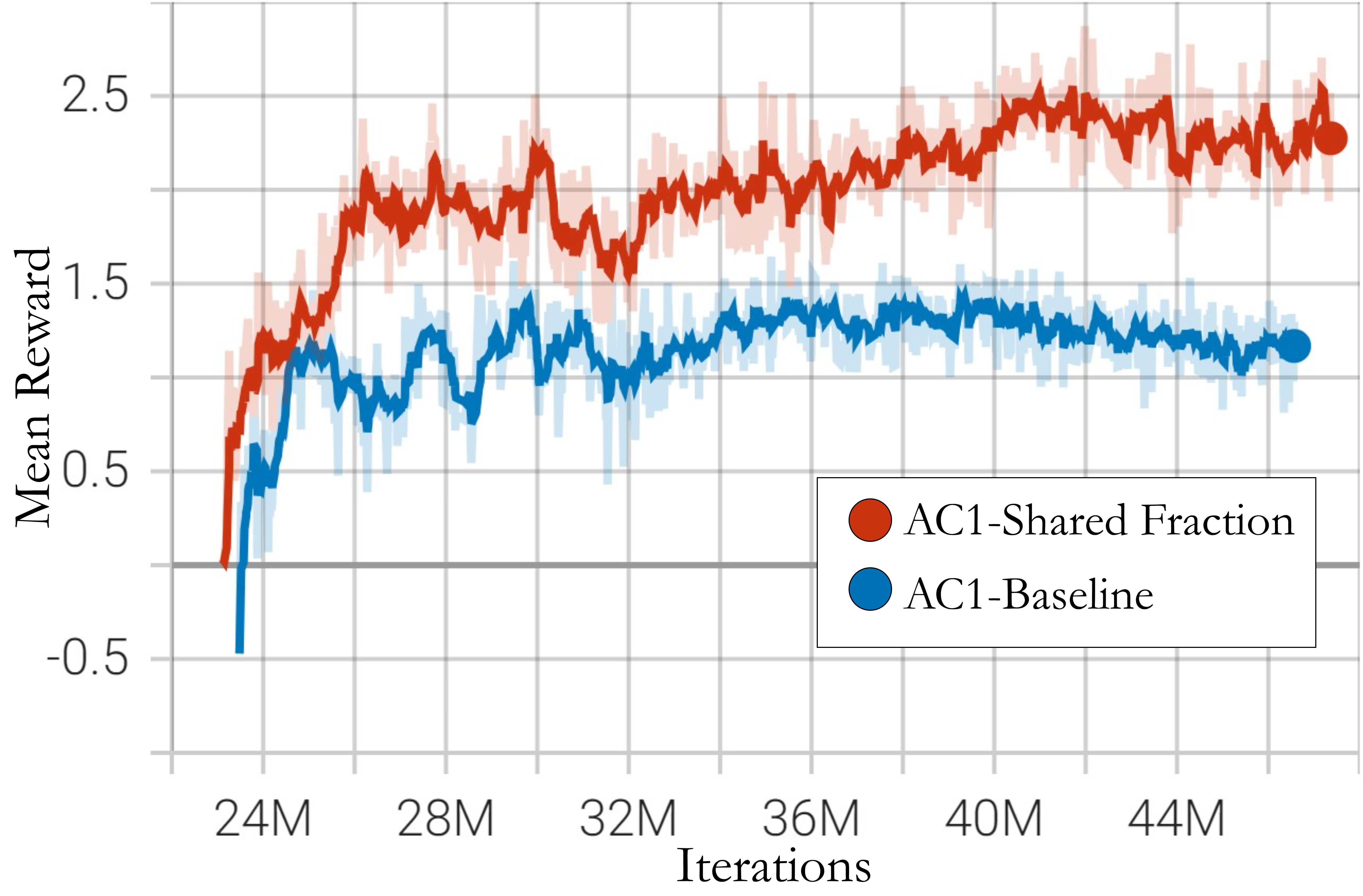}
\label{fig:AC1_rews_comp}}}
\hspace{10pt}
\subfloat[Training comparison of AC2 at L3.]{%
\resizebox*{6cm}{!}{\includegraphics{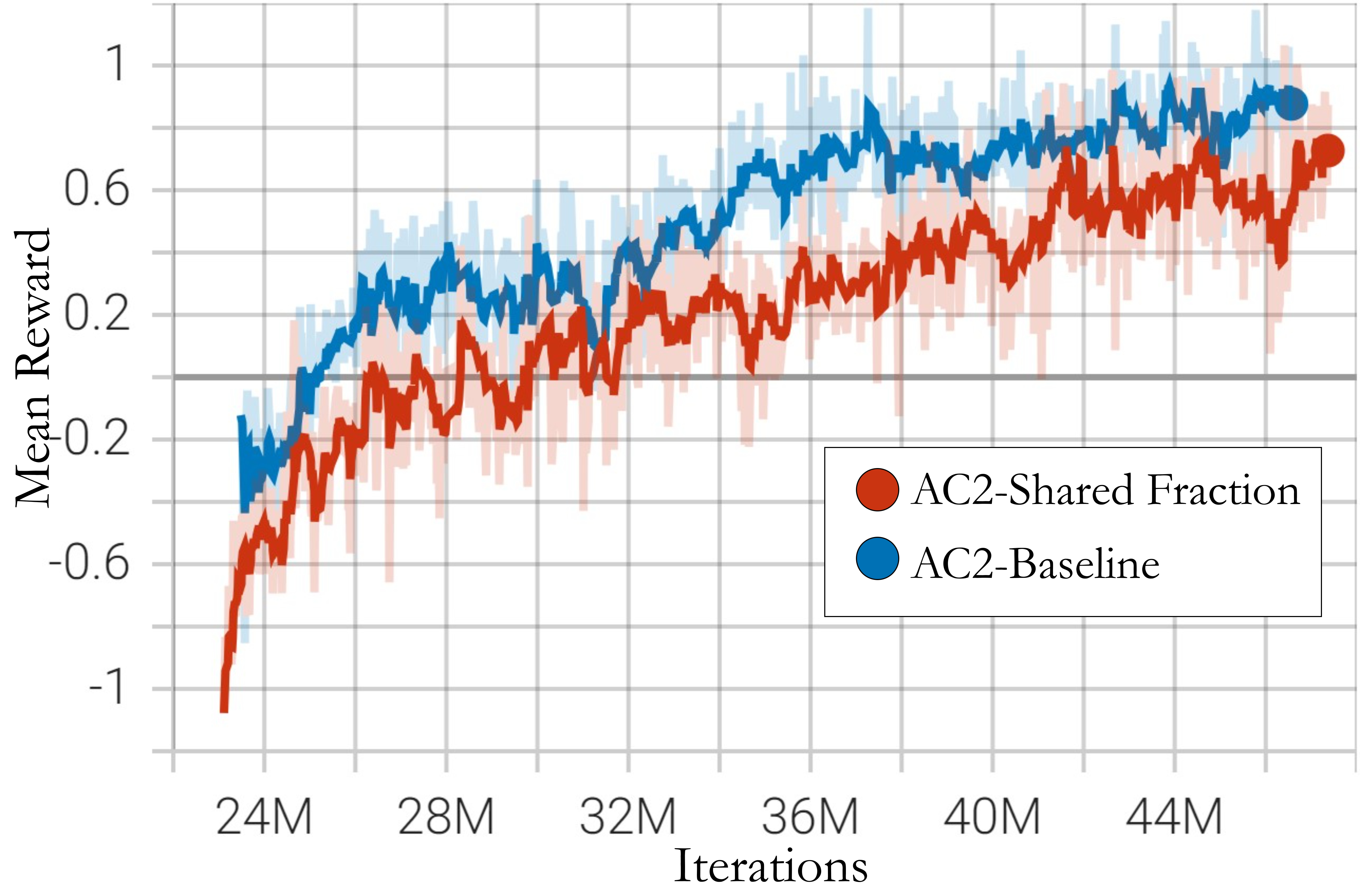}
\label{fig:AC2_rews_comp}}}
\caption{Training of AC1 and AC2 under different fight rewards: individual rewards (Baseline) and a shared reward (Shared Frarction).}
\label{fig:fight_rews_AC1_AC2}
\end{figure}

We delve deeper to understand the underlying causes of the diminished performance observed with reward sharing (\emph{ShFrac}). To achieve this, we analyze and compare the average rewards of agent types AC1 and AC2 under both the base and shared reward mechanisms. As illustrated in Fig.~\ref{fig:AC1_rews_comp}, we observe that AC1 agents perform consistently better under the shared reward condition compared to the performance with baseline reward. Conversely, AC2 exhibits a reduced training performance under shared rewards (Fig.~\ref{fig:AC2_rews_comp}). Whereas both AC1 and AC2 accrue similar average rewards (approximately 1) under the base reward mechanism, yet a notable disparity emerges with shared rewards, indicating the presence of the credit assignment problem due to AC1's superior and AC2's inferior performance. The observed reward mismatch between AC1 and AC2 under \emph{ShFrac} is not desirable, suggesting the potential for lazy agents~\cite{Gronauer2021}. Consequently, we deduce that assigning rewards based solely on an agent's direct combat achievements engenders optimal performance outcomes.

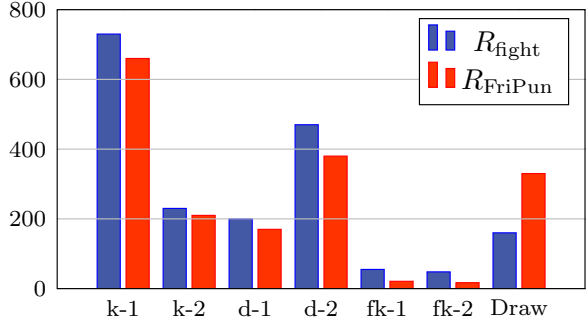
\begin{figure}[htb]
\centering
\resizebox*{8cm}{!}{
\begin{tikzpicture}
\begin{axis}[
ybar,
ymin=0,
ymax=800,
bar width=8,
width=8cm,
ymajorgrids,
height=5cm,
xmin=0,
xmax=8,
yticklabels={{\scriptsize 0}, {\scriptsize 0}, {\scriptsize 200}, {\scriptsize 400}, {\scriptsize 600}, {\scriptsize800}},
x tick label style={align=center,text width=1cm},
xticklabels={{\scriptsize k-1}, {\scriptsize k-2}, {\scriptsize d-1}, {\scriptsize d-2}, {\scriptsize fk-1}, {\scriptsize fk-2}, {\scriptsize Draw},},
xtick={1,2,3,4,5,6,7},
tickwidth=0mm,
axis on top,
legend pos = north east]

\addplot[color=blue, fill=blue!65!lime] coordinates {(1,730) (2,230) (3,200) (4,470) (5,55) (6,48) (7,160)}; \addlegendentry{\small $R_{\mathrm{fight}}$}

\addplot[color=red, fill=red!60!orange] coordinates {(1,660) (2,210) (3,170) (4,380) (5,21) (6,17) (7,330)}; \addlegendentry{\small $R_{\mathrm{FriPun}}$}
\end{axis}
\end{tikzpicture}
\label{plot:XX}}
\caption{Evaluation statistics of $\pi_f$ under $R_{\mathrm{fight}}$ and $R_{\mathrm{FriPun}}$ rewards. The y-axis displays the number of counted events per category. Destroying an opponent is abbreviated with \emph{k}, getting destroyed with \emph{d} and \emph{fk} indicates ``friendly'' kills. The attached numbers indicate the aircraft types, e.g., $k-1$ kill by AC1, $fk-1$ friendly kill by AC1 and $d-1$ means AC1 destroyed.} 
\label{plot:statistic_events_rews_compare}
\end{figure}

To investigate the performance impact of the reward $R_{\mathrm{FriPun}}$, we present detailed statistics of distinct events during evaluation in Fig.~\ref{plot:statistic_events_rews_compare}. These statistics from the 2-vs-2 evaluation scenarios show the different contributions of agents and opponents under both the baseline and \emph{FriPun} reward functions. Overall, the combat skills of AC1 surpass those of AC2, which is most likely due to rockets as further equipment and having more agile dynamics. Upon closer examination of the data, we observe a slight reduction in the frequency of each event under the \emph{FriPun} reward. Although the reduction of friendly fire events is achieved, there was a notable increase in draws, indicating a more cautious behavior, as also evident in Fig.~\ref{fig:fight_rews_compare}. This suggests that accepting the risk of friendly fire can still enhance combat performance, thereby supporting the appropriateness of the reward function $R_{\mathrm{fight}}$.

\begin{figure}[htb]
\centering
    \begin{tikzpicture}
    \begin{axis}[
    xbar stacked,
    legend entries={{\scriptsize Win},{\scriptsize Draw},{\scriptsize Loss}},
    bar width=7,
    width=6cm,
    xmajorgrids,
    height=3.5cm,
    xmin=0,
    xmax=100,
    ytick={0,1,2,3},
    yticklabels={{\footnotesize 2-vs-2},{\footnotesize 3-vs-3}, {\footnotesize 4-vs-4}, {\footnotesize 5-vs-5}},
    xtick={0,50,100},
    tickwidth=0mm,
    legend style={at={(0.5,-0.35)},anchor=north,legend columns=-1},
    xticklabels={{\scriptsize $0\%$},{\scriptsize $50\%$},{\scriptsize $100\%$}},
    axis on top]
    \addplot [color=blue, fill=blue!65!lime] coordinates 
    {(52,0) (53,1) (57,2) (57,3)};
    \addplot [color=gray, fill=gray!60!white] coordinates 
    {(23,0) (22,1) (22,2) (23,3)};
    \addplot [color=red, fill=red!60!orange] coordinates 
    {(25,0) (24,1) (21,2) (20,3)};
    \end{axis}
    \end{tikzpicture}
\caption{Evaluation of $\pi_{f,L5}$-vs-$\pi_{f,L4}$ under $R_{\mathrm{fight}}$ after completing L5 training.}
\label{fig:fight_eval_L5}
\end{figure}
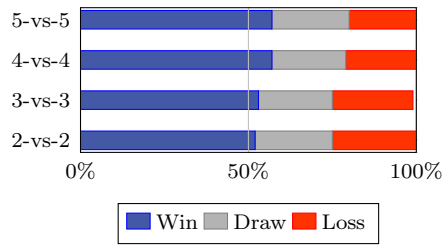

Upon validating the adequacy of our reward function $R_{\mathrm{fight}}$, we assess the performance subsequent to the completion of the entire low-level training phase, with the outcomes depicted in Fig.~\ref{fig:fight_eval_L5}. The training graph of the complete curriculum learning stages is omitted here, as it is illustrated in Fig.~\ref{fig:standard_ctde_train} for the validation of our overall hierarchical approach. To properly evaluate performance, each agent is deployed with $\pi_f$ of L5 and every opponent with $\pi_f$ of L4. We infer that our agents could further improve their combat performance during L5 training stage as $\pi_f$ of L5 outperforms $\pi_f$ of L4. Additionally, our agents are capable of engaging in scenarios up to 5-vs-5, even though training was conducted in a 2-vs-2 scheme. Interestingly, the proportion of combat outcomes remains constant and there is a slight increase in the number of wins as the number of aircraft grows, even though the opponents are better equipped with ammunition. Identifying a direct cause for this improvement is challenging, given that the simulation design ensures the presence of at least one of each aircraft type per team. One plausible explanation for the improved performance with an increasing number of agents could be the enhanced probability of eliminating opponents due to the presence of more L5 agents. For instance, in a 2-vs-2 scenario, if one agent is eliminated, only one remains to combat against two opponents, which results in a higher risk of loosing the battle than in a scenario where one out of five agents is eliminated. From the training graph in Fig.~\ref{fig:standard_ctde_train}, particularly during the L5 stage, where the learning curve does not rise anymore, and due to the solid evaluation performance, we deduce that our agents have achieved their peak learning and combat capabilities. An illustrative example of a combat scenario is presented in Fig.~\ref{fig:traj_fight}, showing the circular trajectories to reach the tail of the opponents.

\subsubsection{Architecture Inspection}\label{sec:fight_architect_inspect}
\begin{figure}[htb]
\centering
\subfloat[Training $\pi_f$ with different architectures.]{%
\resizebox*{6cm}{!}{\includegraphics{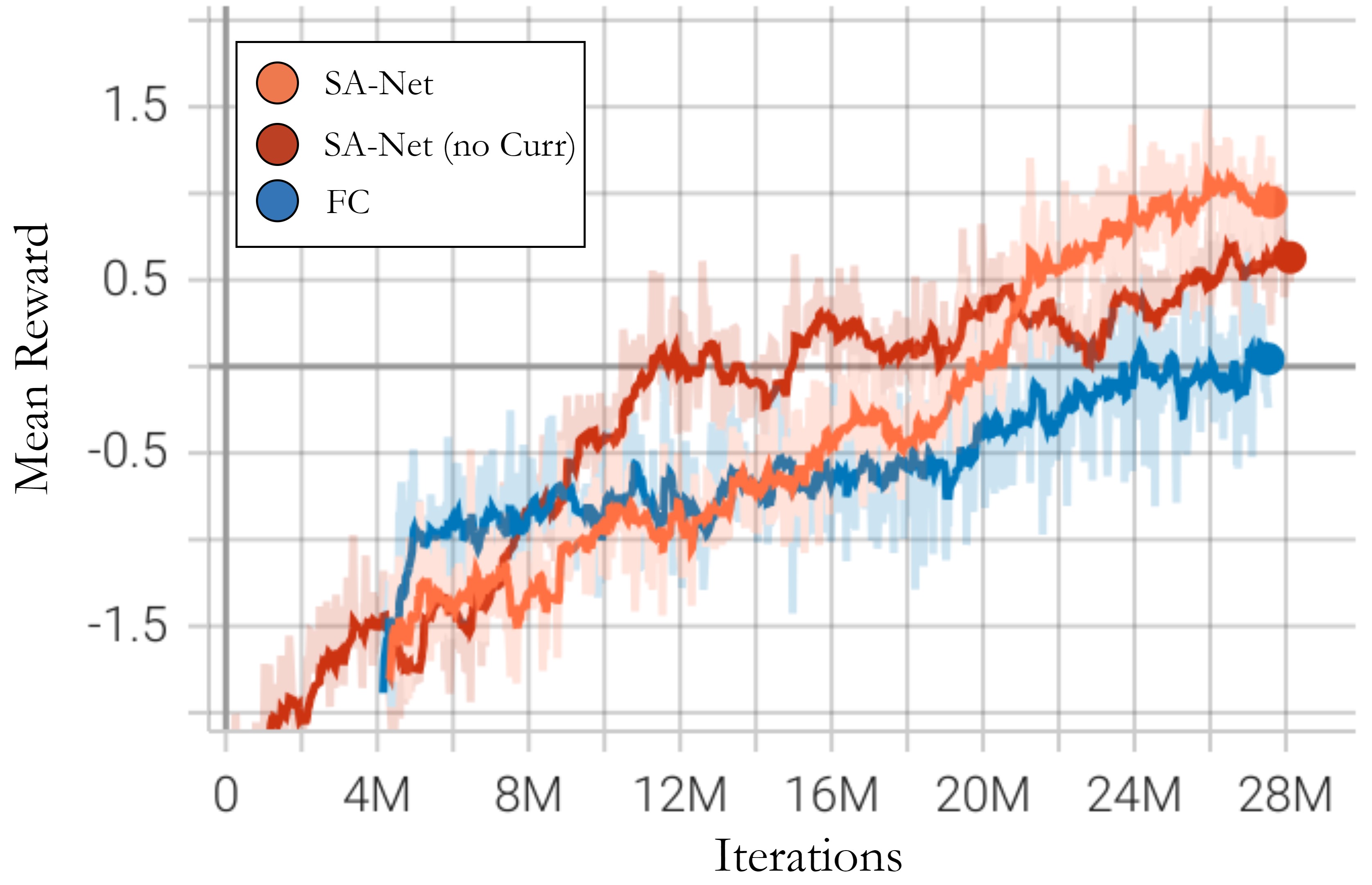}
\label{fig:fight_arch_train}}}
\hspace{5pt}
\subfloat[Evaluation $\pi_f$ with different architectures.]{%
\resizebox*{6cm}{!}{
\begin{tikzpicture}
\begin{axis}[
xbar stacked,
legend entries={{\scriptsize Win},{\scriptsize Draw},{\scriptsize Loss}},
bar width=8,
width=5cm,
xmajorgrids,
height=3cm,
xmin=0,
xmax=100,
ytick={0,1,2},
yticklabels={{\footnotesize FC},{\footnotesize SA-Net (no Curr)}, {\footnotesize SA-Net}},
xtick={0,50,100},
tickwidth=0mm,
legend style={at={(0.5,-0.35)},anchor=north,legend columns=-1},
xticklabels={{\scriptsize $0\%$},{\scriptsize $50\%$},{\scriptsize $100\%$}},
axis on top]
\addplot [color=blue, fill=blue!65!lime] coordinates
{(39,0) (53,1) (67,2)};
\addplot [color=gray, fill=gray!60!white] coordinates
{(28,0) (30,1) (19,2)};
\addplot [color=red, fill=red!60!orange] coordinates
{(33,0) (17,1) (14,2)};
\end{axis}
\end{tikzpicture}
\label{plot:fight_arch_eval}}}
\caption{Training performance of $\pi_f$ under different architectures.} 
\label{fig:fight_arch_compare}
\end{figure}
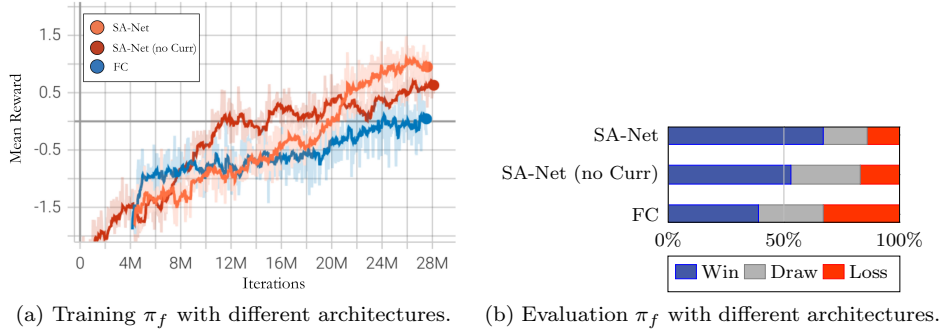

To underscore the effectiveness of our network architecture and the influence of curriculum learning, we train $\pi_f$ using various architectures and compare their performances as depicted in Fig.~\ref{fig:fight_arch_compare}. We take a Fully-Connected (FC) neural network to compare against our presented SA-Net (as defined in Sec.~\ref{sec:overall_structure}). This FC-Net comprises two layers with \num{500} neurons each and uses $tanh$ activation. Additionally, we conducted a training experiment the SA-Net without the curriculum stages, opting instead for direct training on L3. Notably, the results indicate a reasonable enhancement in both training and evaluation performance with our proposed methodology.

\subsubsection{MARL Framework Inspection}\label{sec:marl_framework_inspect}
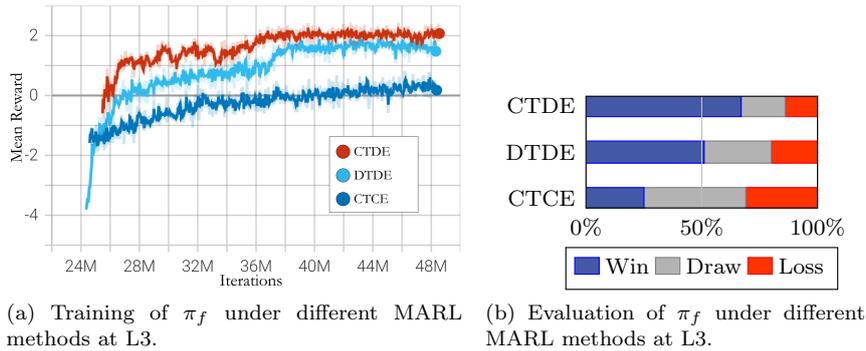
\begin{figure}[htb]
\centering
\subfloat[Training of $\pi_f$ under different MARL methods at L3.]{%
\resizebox*{6cm}{!}{\includegraphics{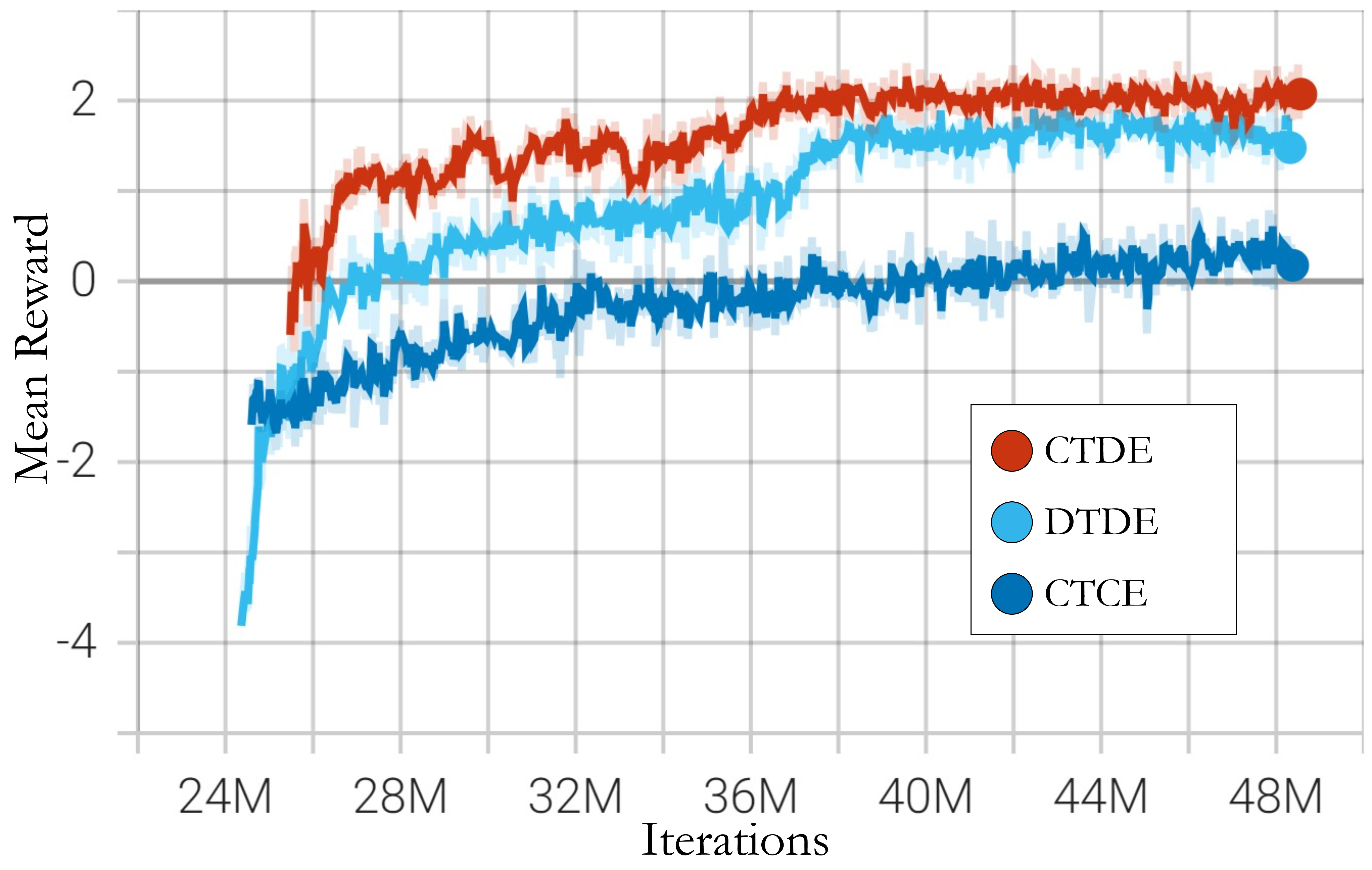}
\label{fig:marl_algos_rews}}}
\hspace{5pt}
\subfloat[Evaluation of $\pi_f$ under different MARL methods at L3.]{%
\resizebox*{5cm}{!}{
\begin{tikzpicture}
\begin{axis}[
xbar stacked,
legend entries={{\scriptsize Win},{\scriptsize Draw},{\scriptsize Loss}},
bar width=8,
width=4.5cm,
xmajorgrids,
height=3cm,
xmin=0,
xmax=100,
ytick={0,1,2},
yticklabels={{\footnotesize CTCE},{\footnotesize DTDE}, {\footnotesize CTDE}},
xtick={0,50,100},
tickwidth=0mm,
legend style={at={(0.5,-0.35)},anchor=north,legend columns=-1},
xticklabels={{\scriptsize $0\%$},{\scriptsize $50\%$},{\scriptsize $100\%$}},
axis on top]
\addplot [color=blue, fill=blue!65!lime] coordinates
{(25,0) (51,1) (67,2)};
\addplot [color=gray, fill=gray!60!white] coordinates
{(44,0) (29,1) (19,2)};
\addplot [color=red, fill=red!60!orange] coordinates
{(31,0) (20,1) (14,2)};
\end{axis}
\end{tikzpicture}
\label{plot:marl_algos_eval}}}
\caption{Performance comparison of different MARL frameworks.} 
\label{marl_algos_compare}
\end{figure}

In the realm of MARL, there are two further principal training paradigms besides CTDE, that is already introduced in Sec.~\ref{sec:MARL_def}: \emph{Centralized Training with Centralized Execution} (CTCE) and \emph{Decentralized Training with Decentralized Execution} (DTDE), with CTDE being the primary choice for training in the multi-agent domain~\cite{Gronauer2021}. Nevertheless, the superiority of CTDE is not universally applicable~\cite{ctde_inspection}. This leads us to thoroughly examine which framework yields the best performance in our specific context. Within the CTDE paradigm, we employ shared networks for agents of same aircraft type. Conversely, in the DTDE approach, each agent operates its distinct network. Meanwhile, the CTCE strategy integrates a single network to process the aggregated observations and actions of all agents. Through training and assessing each model in a 3-vs-3 aerial combat scenario, our findings are depicted in Fig.~\ref{marl_algos_compare}. From results we infer, that CTCE trails in both training and evaluative measures. DTDE showcases a solid training performance, similar to CTDE, but falls short during evaluation. Despite the effectiveness of the CTDE framework, a minor proportion of losses persist during evaluation. These losses predominantly occur in situations where the agent and its opponent are directly facing each other and simultaneously firing, thereby creating a scenario where both have an equal likelihood of eliminating the other. Nonetheless, our analysis confirms that CTDE emerges as the most proficient framework in our examination.

\subsection{Escape Policy $\pi_e$}\label{sec:esc_policy_results}
The training process of $\pi_e$ is configured similarly to that of $\pi_f$, utilizing the same ammunition and time horizons. However, the curriculum stages are omitted, and training is conducted directly on L3. The investigation focuses on the evading capabilities under various training configurations.

\subsubsection{Reward Inspection}\label{sec:esc_rew_inspect}
In escape mode, the agents' main objective is to evade opponent aircraft, making a distance-based reward function a logical choice to encourage maintaining maximum distance from adversaries. We therefore compare the training reward defined in Eq.~\eqref{eq:rew_esc} with two distinct reward functions in escape mode:
\begin{equation}
    R_{\mathrm{dist},t} = R_{\mathrm{esc}} + R_{p,t}\,,
\label{eq:esc_dist}
\end{equation}
\begin{equation}
    R_{\mathrm{dist-speed},t} = R_{\mathrm{esc}} + R_{pv,t}\,.
\label{eq:esc_dist_speed}
\end{equation}

Each reward function is evaluated individually (i.e., per agent) and includes penalties for boundary violations ($R_b$), being destroyed ($R_d$), and friendly kills ($R_{fr}$), rendering the baseline function $R_{\mathrm{esc}}$ strictly non-positive, with the primary goal to avoid penalties. The introduced reward functions in Eq.~\eqref{eq:esc_dist} and~\eqref{eq:esc_dist_speed} further include two per-time-step rewards, where $d_o$ is the distance to opponents and $v_{\mathrm{AC}}$ is the velocity as defined in Table~\ref{table:aircraft_dynamics}:

\begin{align}
     R_{p,t} &=
    \begin{cases}
    \num{-0.1} & d_o<\qty{6}{\kilo\metre}\\
    +\num{0.1} & d_o>\qty{13}{\kilo\metre} \\
    0 & \mathrm{otherwise},
    \end{cases} 
    \label{eq:esc_p} \\
     R_{pv,t} &=
    \begin{cases}
    \num{-0.1} & d_o<\qty{6}{\kilo\metre} \wedge v_{\mathrm{AC}}<\qty{300}{\knot}\\
    +\num{0.1} & d_o>\qty{13}{\kilo\metre} \wedge v_{\mathrm{AC}}>\qty{600}{\knot}\\
    0 & \mathrm{otherwise}.
    \end{cases} \label{eq:esc_pv}
\end{align}

\begin{figure}[htb]
\centering
\subfloat[Training of $\pi_e$ at L3.]{%
\resizebox*{6cm}{!}{\includegraphics{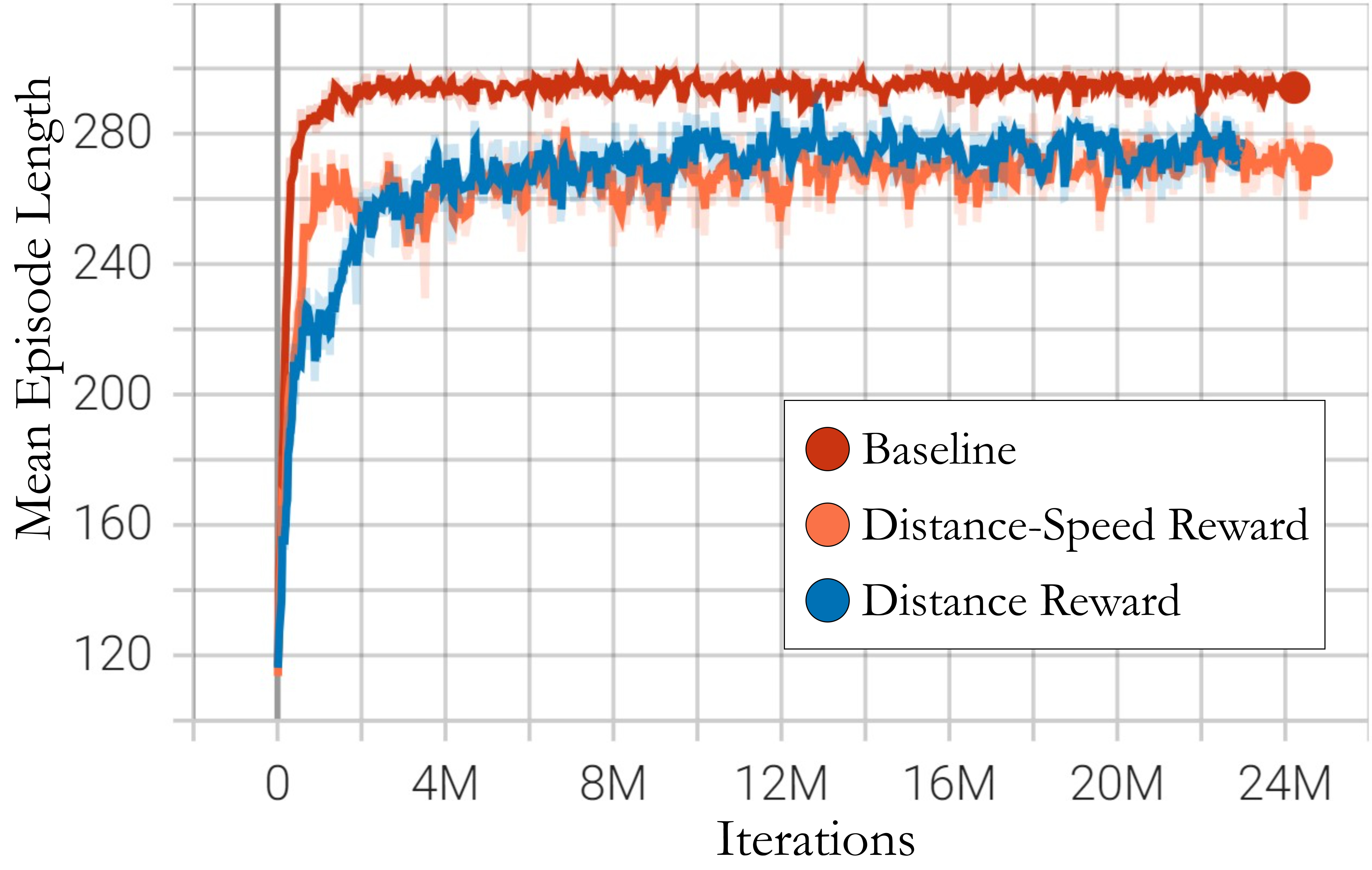}
\label{fig:esc_rews_train}}}
\hspace{5pt}
\subfloat[Evaluation $\pi_e$ against script opponents.]{%
\resizebox*{6cm}{!}{
\begin{tikzpicture}
\begin{axis}[
xbar stacked,
legend entries={{\scriptsize Escaped},{\scriptsize Kills},{\scriptsize Killed}},
bar width=8,
width=5cm,
xmajorgrids,
height=3cm,
xmin=0,
xmax=100,
ytick={0,1,2},
yticklabels={{\footnotesize Dist},{\footnotesize Dist-Speed}, {\footnotesize Baseline}},
xtick={0,50,100},
tickwidth=0mm,
legend style={at={(0.5,-0.35)},anchor=north,legend columns=-1},
xticklabels={{\scriptsize $0\%$},{\scriptsize $50\%$},{\scriptsize $100\%$}},
axis on top]
\addplot [color=blue, fill=blue!65!lime] coordinates
{(66,0) (61,1) (72,2)};
\addplot [color=gray, fill=gray!60!white] coordinates
{(15,0) (12,1) (19,2)};
\addplot [color=red, fill=red!60!orange] coordinates
{(19,0) (27,1) (9,2)};
\end{axis}
\end{tikzpicture}
\label{plot:esc_rews_eval}}}
\caption{Comparison of performance under different escape rewards. \emph{Escaped} means no agent got killed (including going out of boundary), \emph{killed} is when at least one agent got killed, \emph{kills} when at least one opponent got killed.} 
\label{fig:esc_eplen_compare}
\end{figure}
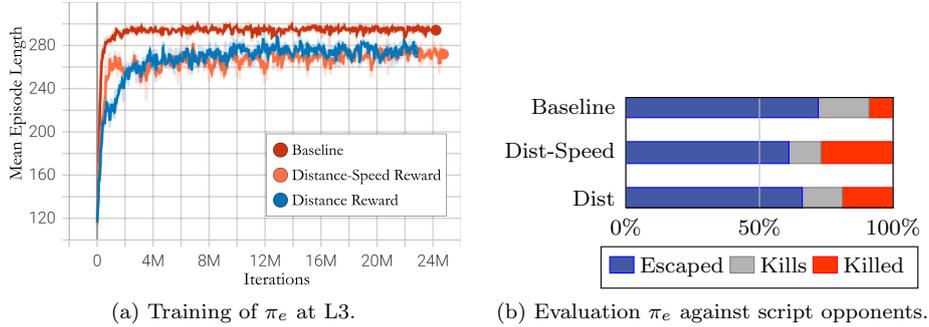

The functions in Equations~\eqref{eq:esc_p} and~\eqref{eq:esc_pv} are designed to incentivize evasion by rewarding agents for maintaining a significant distance from opponents and avoid proximity, with $R_{pv,t}$ further taking into account the agent's velocity, underpinning the intuition that a higher speed decreases the likelihood of getting caught. Rather than focusing on reward values, we analyze mean episode lengths during training as an indicator of agent survival duration. With a maximum episode length of \num{300} time-steps for L3 training, the data illustrates that agents trained with $R_{\mathrm{esc}}$ exhibit the longest survival times (Fig.~\ref{fig:esc_rews_train}). Both per-time-step rewards yield similar, albeit slightly inferior, performance to the baseline function. Evaluation stages reveal an interesting increase in number of defeats, especially when trained with $R_{\mathrm{dist-speed},t}$ (Fig.~\ref{plot:esc_rews_eval}). Further examination of rendered scenarios indicates that with the increased velocity, agents tend to exit the designated area more frequently. Consequently, training with per-time-step rewards appears less effective, with the penalties for eliminations being less impactful in the learning process. This occurs because agents can accumulate sufficient positive rewards during an episode such that, even with the penalty for being destroyed, the total reward may still remain positive. Therefore, we conclude that $R_{\mathrm{esc}}$ reveals the most effective escaping performance.

\subsubsection{Fleeing Performance}\label{sec:esc_L5_performance}

\begin{figure}[htb]
\centering
\subfloat[Training of $\pi_e$-vs-$\pi_{f,L5}$.]{%
\resizebox*{6cm}{!}{\includegraphics{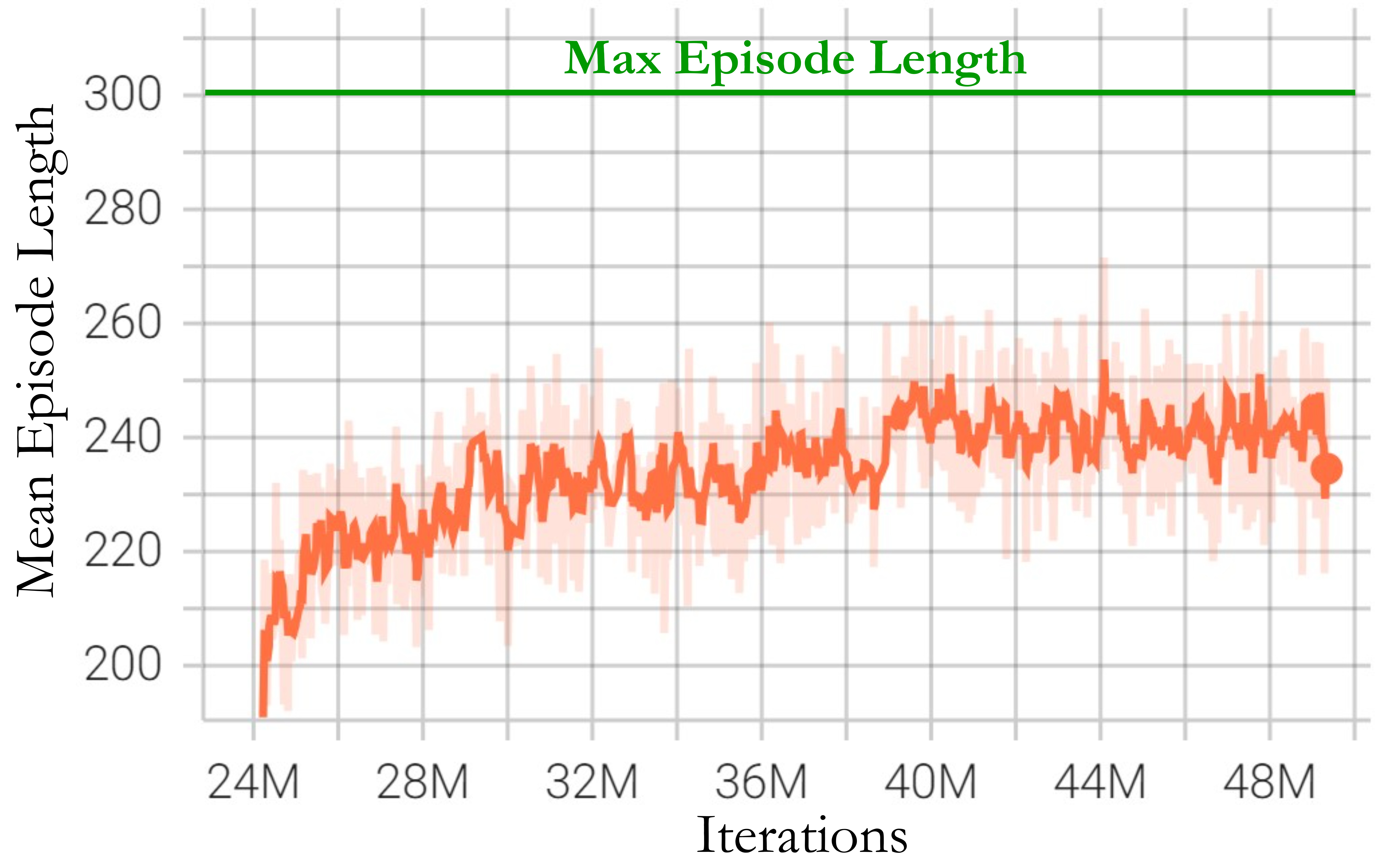}
\label{fig:esc_rews_L5_eplen}}}
\hspace{5pt}
\subfloat[Evaluation of $\pi_e$-vs-$\pi_{f,L5}$.]{%
\resizebox*{5.5cm}{!}{
\begin{tikzpicture}
\begin{axis}[
xbar stacked,
legend entries={{\scriptsize Escaped},{\scriptsize Kills},{\scriptsize Killed}},
bar width=7,
width=5cm,
xmajorgrids,
height=3cm,
xmin=0,
xmax=100,
ytick={0,1,2,3},
yticklabels={{\footnotesize 2-vs-2},{\footnotesize 3-vs-3}, {\footnotesize 4-vs-4}, {\footnotesize 5-vs-5}},
xtick={0,50,100},
tickwidth=0mm,
legend style={at={(0.5,-0.35)},anchor=north,legend columns=-1},
xticklabels={{\scriptsize $0\%$},{\scriptsize $50\%$},{\scriptsize $100\%$}},
axis on top]
\addplot [color=blue, fill=blue!65!lime] coordinates 
{(79,0) (66,1) (49,2) (37,3)};
\addplot [color=gray, fill=gray!60!white] coordinates 
{(5,0) (7,1) (13,2) (14,3)};
\addplot [color=red, fill=red!60!orange] coordinates 
{(16,0) (27,1) (38,2) (49,3)};
\end{axis}
\end{tikzpicture}
\label{plot:esc_eval_L5}}}
\caption{Performance of $\pi_e$ at against $\pi_{f,L5}$. \emph{Escaped} means no agent got killed (including going out of boundary), \emph{killed} is when at least one agent got killed, \emph{kills} when at least one opponent got killed.} 
\label{fig:esc_L5}
\end{figure}
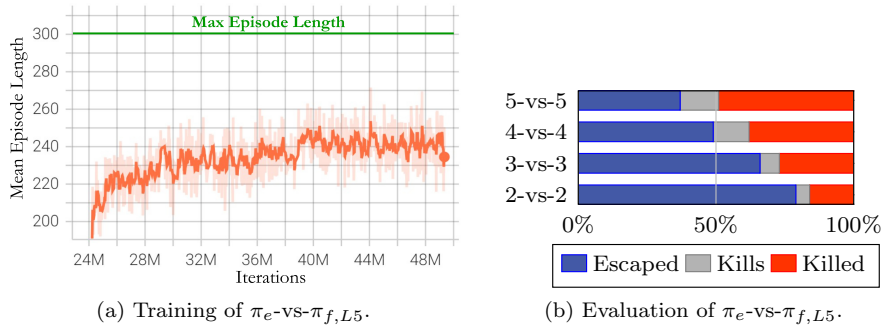

Upon confirming the appropriate choice of our selected reward function for training $\pi_e$, we proceeded with the training process by competing against $\pi_f$ from L5, which possesses the most advanced combat skills. We maintained the time horizon $T=\num{300}$, as during L3 training phase. Subsequently, we plot the mean episode length and assessed the performance of $\pi_e$ across various combat scenarios, as depicted in Fig.~\ref{fig:esc_L5}. The training results reveal that the fleeing agents did not come as close to the maximum episode length as during L3 training. Nevertheless, with successful evasion in over \qty{50}{\percent} of the encounters in 3-vs-3 scenarios and below, the performance of $\pi_e$ remains appealing, as demonstrated in the evaluation plot. With increasing number of aircraft involved, the likelihood of an agent being killed also rises. Nevertheless, a significant number of successful evasions are observed, even in more complex combat scenarios. An example of fleeing trajectories are plotted in Fig.~\ref{fig:traj_esc}.

\subsection{Commander Policy $\pi_c$}\label{sec:commandeR_policy_results}

The objective of the commander policy $\pi_c$ is to provide strategic commands (either to attack or to escape) to the low-level policies. Depending on the command issued by $\pi_c$, agents are assigned either $\pi_f$ from L5 or $\pi_e$. Opponents predominantly engage in combat, with a minor propensity to flee, possessing a fight-to-escape ratio of approximately \num{3}:\num{1} ($p_o=\qty{75}{\percent}$ in Algorithm~\ref{hieR_algo}). Compared to the training of $\pi_f$ and $\pi_e$, we equip agents and opponents with an equal arsenal of \num{300} cannons and \num{8} rockets. This setup facilitates a focused comparison of tactical approaches: agents employing commander tactics opponents utilizing random tactics.

\subsubsection{Commander Modifications}\label{sec:hieR_modifications}
The base definition of the commander structure is defined in Sec.~\ref{sec:commandeR_policy}. However, we are interested to inspect the impact of some fundamental changes in the hierarchical structure. We define the following main modifications:

\begin{itemize}
    \item \emph{Shared-vs-Global.} Comparison of a shared CTDE Network for each agent versus a global CTCE Network for all agents. As in the low-level training with the CTCE approach, the observations and actions are combined, as detailed in Sec.~\ref{sec:marl_framework_inspect}.
    \item \emph{N2-vs-N3.} We evaluate varying sensing capabilities. In the N2 configuration, the commander can detect the two nearest opponents for each agent, while in the N3 setting, the commander senses three opponents. Correspondingly, the commander's action space is modified to choose from among the two or three opponents. Training is conducted in a 3-vs-3 combat scenario in both configurations.
    \item \emph{Opt-vs-noOpt.} In the basic definition, the commander has the capability to select \emph{which} opponent to attack or to initiate escape mode for the agent. This is termed the \emph{Opt} behavior. We contrast this with a scenario where the commander can \emph{only} decide whether to attack or flee for each agent, referred to as \emph{noOpt}. When the attack option is chosen, the low-level policy targets the nearest opponent. 
    \item \emph{Assess-vs-noAssess.} The term \emph{Assess} describes the training of the commander, incorporating the intrinsic reward $R_{\mathrm{act}}$, as outlined in Sec.~\ref{sec:commandeR_policy}. We evaluate the commander's performance without this additional reward, which we label as \emph{noAssess}. In this scenario, the commander receives the raw combat rewards: $R_k + R_d + R_b$.
\end{itemize}

We trained the commander using all combinations of these modifications. The evaluation performance is depicted in Table~\ref{table:hieR_configs}. The proportion of attack and escape commands is presented in the right column. The top-performing approaches in both shared and global frameworks are emphasized in bold. We confirm that our method, detailed in Sec.~\ref{sec:commandeR_policy}, achieves the highest combat effectiveness, closely followed by the CTCE framework approach. Further analysis reveals that the N2 configurations consistently outperform their N3 counterparts by approximately a \qty{10}{\percent} win margin with the CTDE framework, while the differences are less pronounced with the CTCE framework. As a result of the enhanced possibilities of CoAs, the \emph{Opt} setting improves performance across all configurations. The action assessment reward $R_{\mathrm{act}}$ exhibits varying effects. While training the commander under CTDE with the $R_{\mathrm{act}}$ generally resulted in improved performance, the opposite occurred under the CTCE framework. In this case, the intrinsic reward did not enhance the commander's tactical learning. An additional observation from incorporating $R_{\mathrm{act}}$ during commander training is the alteration in the fight-escape selection ratio. The most significant shifts occurred in the \emph{Shared-N3-Opt} and \emph{Glob-N3-Opt} configurations, with fight ratios exceeding \qty{90}{\percent}. In these cases, the shared and global frameworks exhibited contrasting effects. These two observations regarding the action assessment function $R_{\mathrm{act}}$ may be due to the commander having access to the full environment state, which could simplify the strategic planning process. However, the CTCE approach restricts flexibility as it can only be applied to specific combat settings (otherwise, observations would require inefficient zero-padding).

\begin{table}[htb]
\tbl{Hierarchical configurations and results. All numerical values are given in [\unit{\percent}].}
{\begin{tabular}{m{4cm}m{1.4cm}m{1.4cm}m{1.4cm}m{1.6cm}}
\hline
\textbf{Configuration} & \textbf{Win} & \textbf{Loss} & \textbf{Draw} & \textbf{Fight/Esc} \\ \hline 
\textbf{Shared-N2-Opt-Assess} & \textbf{63.7} & \textbf{16.2} & \textbf{20.1} & 87.1 / 12.9 \\
Shared-N2-Opt-noAssess & 55.9 & 20.2 & 23.9 & 82.2 / 17.8 \\
Shared-N2-noOpt-Assess & 53.5 & 18.4 & 28.1 & 71.1 / 28.9 \\
Shared-N2-noOpt-noAssess & 48.9 & 21.1 & 30.0 & 69.3 / 30.7 \\
Shared-N3-Opt-Assess & 48.2 & 26.3 & 25.5 & 84.9 / 15.1 \\
Shared-N3-Opt-noAssess & 47.4 & 26.7 & 25.9 & 91.2 / 8.8 \\
Shared-N3-noOpt-Assess & 43.3 & 28.6 & 28.1 & 76.5 / 23.5 \\
Shared-N3-noOpt-noAssess & 42.7 & 28.2 & 29.1 & 75.9 / 24.1 \\
Glob-N2-Opt-Assess & 59.1 & 16.5 & 24.4 & 88.9 / 11.1 \\
\textbf{Glob-N2-Opt-noAssess} & \textbf{60.6} & \textbf{16.6} & \textbf{22.8} & 82.1 / 17.9 \\
Glob-N2-noOpt-Assess & 53.4 & 18.2 & 28.4 & 78.7 / 21.3 \\
Glob-N2-noOpt-noAssess & 54.3 & 17.9 & 27.8 & 77.3 / 22.7 \\
Glob-N3-Opt-Assess & 59.9 & 16.9 & 23.2 & 93.7 / 6.3 \\
Glob-N3-Opt-noAssess & 55.5 & 18.8 & 25.7 & 84.8 / 15.2 \\
Glob-N3-noOpt-Assess & 48.8 & 17.4 & 33.8 & 72.3 / 27.7 \\
Glob-N3-noOpt-noAssess & 42.7 & 19.1 & 38.2 & 68.4 / 31.6 \\ \hline
\end{tabular}}
\label{table:hieR_configs}
\end{table}

\subsubsection{Reward Inspection}\label{sec:hieR_rew_inspect}
\begin{figure}[htb]
\centerline{\includegraphics[scale=0.02]{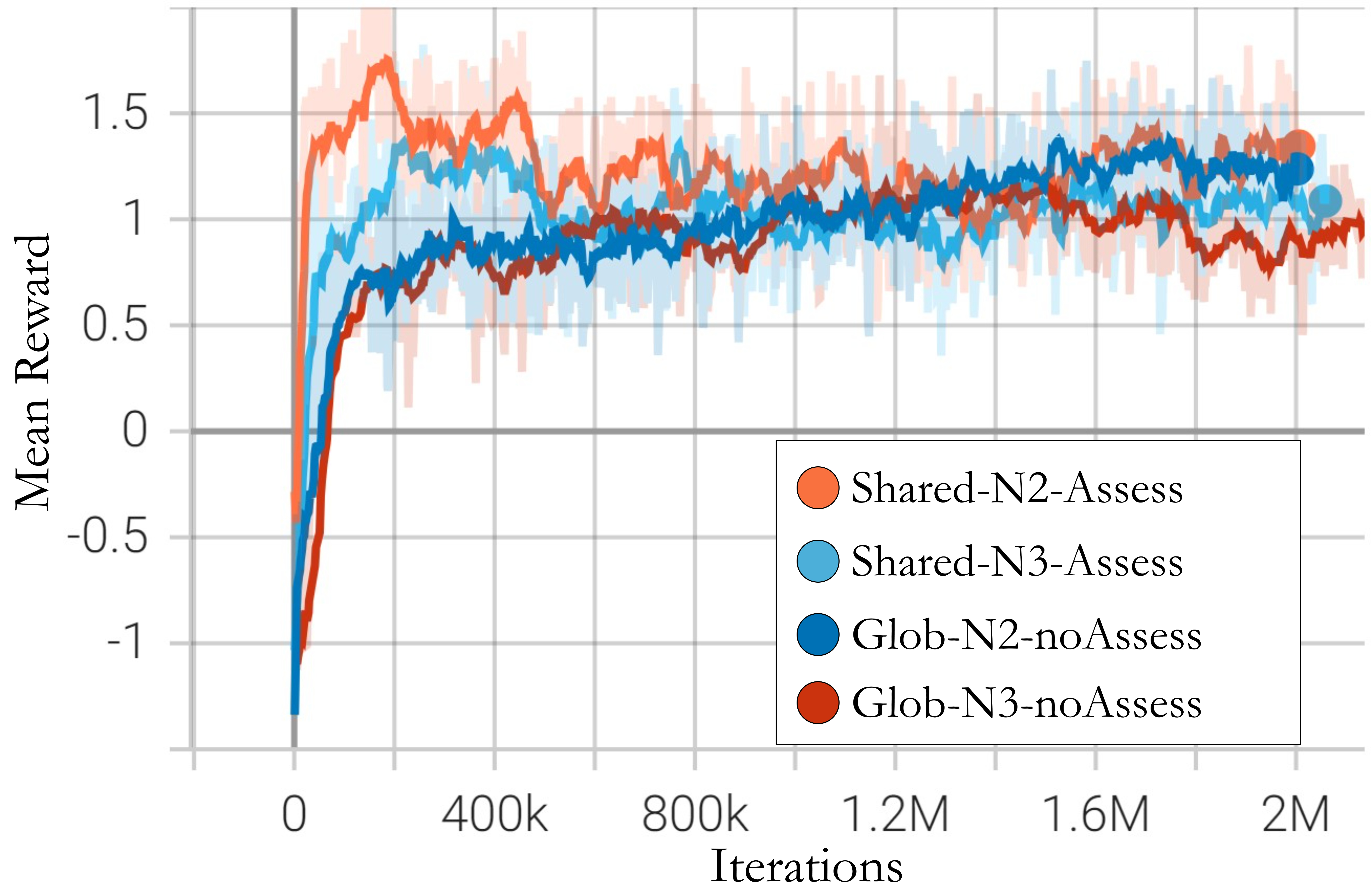}}
\caption{Comparison of the training rewards of the best performing hierarchy configurations.}
\label{fig:best_hieR_rew_comp}
\end{figure}

Upon analyzing various hierarchical structures and commands delineated in Sec.~\ref{sec:hieR_modifications}, we select the optimal shared and global commander policies, as highlighted in Table~\ref{table:hieR_configs}. We particularly examine their N2 and N3 configurations, since these modifications had the most significant impact on performance. Notably, we focus on the \emph{Opt} setting, as it revealed the best performance across all configurations. We inspect the training outcomes of the four configurations with the option selection enabled, depicted in Fig.~\ref{fig:best_hieR_rew_comp}. In the results, the assessment reward $R_{\mathrm{act}}$ is subtracted in the mean reward for the CTDE approach to have a consistent comparison. Overall, we observe a rapid convergence of the mean reward curve. During the initial training phase, there is a considerable difference in reward distributions, with the \emph{Shared-N2-Assess} method exhibiting superiority and highest efficiency. However, towards the latter part of the training process, all four methods converge around a mean reward value of 1, with the N2 configurations slightly outperforming the N3 setting. This observation aligns with the results presented in Table~\ref{table:hieR_configs}. The superior performance of the N2 configurations may be attributed to the reduced complexity of the commander's task and a focused strategy targeting the two nearest opponents.

\subsubsection{Opponent Selection}\label{sec:hieR_opponent_selection}
\begin{figure}[htb]
\centering
\begin{tikzpicture}
\begin{axis}[
xbar stacked,
legend entries={{\scriptsize Opp1},{\scriptsize Opp2},{\scriptsize Opp3}},
bar width=7.5,
width=6cm,
xmajorgrids,
height=3.3cm,
xmin=0,
xmax=100,
ytick={0,1,2,3},
yticklabels={{\footnotesize Shared-N2-Assess},{\footnotesize Shared-N3-Assess}, {\footnotesize Glob-N2-NoAssess}, {\footnotesize Glob-N3-NoAssess}},
xtick={0,50,100},
tickwidth=0mm,
legend style={at={(0.5,-0.35)},anchor=north,legend columns=-1},
xticklabels={{\scriptsize $0\%$},{\scriptsize $50\%$},{\scriptsize $100\%$}},
axis on top]
\addplot [color=blue, fill=blue!65!lime] coordinates 
{(82,0) (68,1) (92,2) (74,3)};
\addplot [color=red, fill=red!60!orange] coordinates 
{(18,0) (30,1) (8,2) (16,3)};
\addplot [color=green, fill=green!70!black] coordinates 
{(0,0) (2,1) (0,2) (10,3)};
\end{axis}
\end{tikzpicture}
\caption{Opponent selection frequency based on best performing hierarchy configurations. Opponent naming is done based on proximity to agents, where Opp1 is the closest and Opp3 the furthest.}
\label{plot:opp_selection_hieR_configs}
\end{figure}
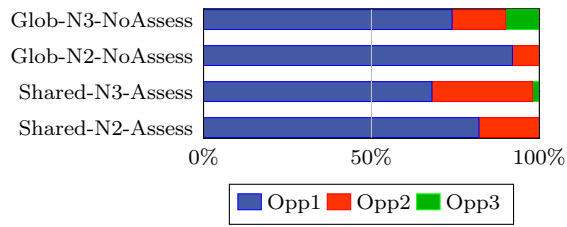

Based on the four commander configurations denoted in Sec.~\ref{sec:hieR_rew_inspect}, we subsequently investigate the opponent selection strategies, again focusing on the N2-vs-N3 configurations to evaluate their sensing capabilities. The assessment outcomes in 3-vs-3 scenarios are depicted in Fig.~\ref{plot:opp_selection_hieR_configs}. Opponents are named according to their distance, with Opp1 being the nearest and Opp3 the furthest. It is evident that under the N2 configuration, the selection of the third opponent is consistently precluded. Predominantly, the nearest opponent (Opp1) is selected with the highest frequency. Nonetheless, a significant selection frequency is also observed for Opp2, especially for \emph{Shared-N3-Assess}. Opp3 is seldom chosen, except in the case of the \emph{Glob-N3-NoAssess} strategy, which appears to favor the most distant opponent more frequently than other strategies. Contrary, \emph{Glob-N2-NoAssess} indicates the most deterministic CoA, focusing mainly the closest enemy aircraft. The variability in the outcomes of the commander policies is challenging to elucidate due to several factors. Primarily, these outcomes are contingent upon the specific combat scenarios, as well as the dynamics of the low-level policies during the intervals between successive commander activations, as outlined in Algorithm~\ref{hieR_algo}. This complexity arises from the interplay between strategic decisions made at the commander level and the low-level executions, which collectively influence the efficacy and the tactical behavior of the commander policy. Overall, these findings suggest that allowing the commander to autonomously determine the optimal tactics by deciding which opponent to attack enhances performance.

\subsubsection{Architecture Inspection}\label{sec:hieR_architect_inspect}
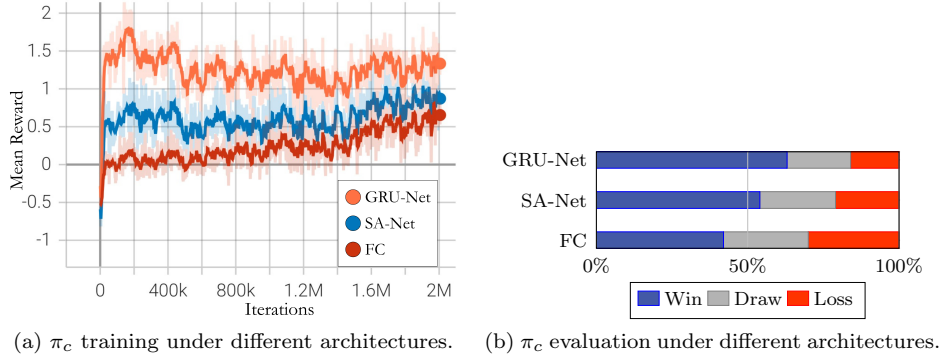
\begin{figure}[htb]
\centering
\subfloat[$\pi_c$ training under different architectures.]{%
\resizebox*{6cm}{!}{\includegraphics{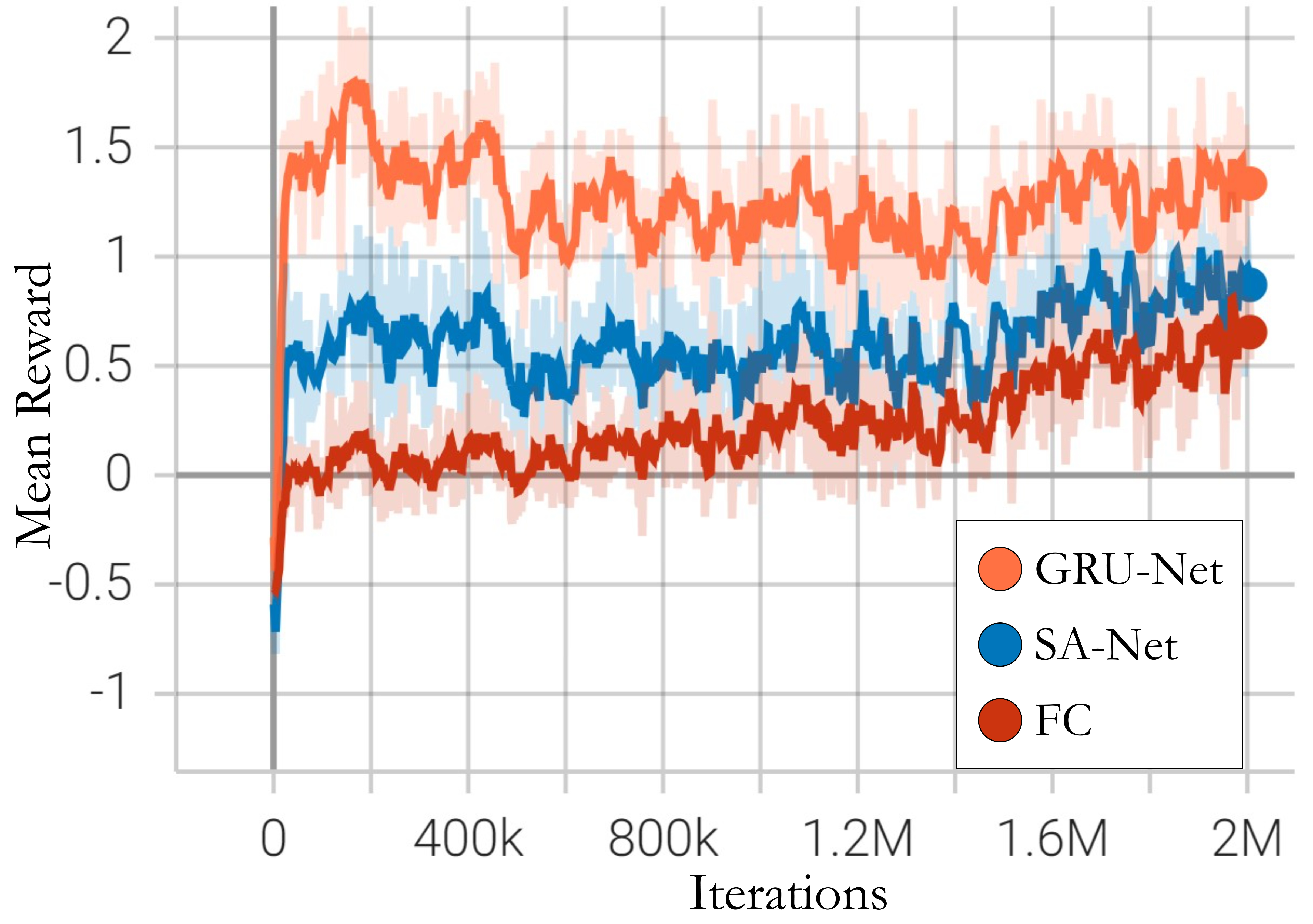}
\label{fig:hieR_rew_arch}}}
\hspace{5pt}
\subfloat[$\pi_c$ evaluation under different architectures.]{%
\resizebox*{6cm}{!}{
\begin{tikzpicture}
\begin{axis}[
xbar stacked,
legend entries={{\scriptsize Win},{\scriptsize Draw},{\scriptsize Loss}},
bar width=7,
width=6cm,
xmajorgrids,
height=3cm,
xmin=0,
xmax=100,
ytick={0,1,2},
yticklabels={{\footnotesize FC},{\footnotesize SA-Net}, {\footnotesize GRU-Net}},
xtick={0,50,100},
tickwidth=0mm,
legend style={at={(0.5,-0.35)},anchor=north,legend columns=-1},
xticklabels={{\scriptsize $0\%$},{\scriptsize $50\%$},{\scriptsize $100\%$}},
axis on top]
\addplot [color=blue, fill=blue!65!lime] coordinates 
{(42,0) (54,1) (63,2)};
\addplot [color=gray, fill=gray!60!white] coordinates 
{(28,0) (25,1) (21,2)};
\addplot [color=red, fill=red!60!orange] coordinates 
{(30,0) (21,1) (16,2)};
\end{axis}
\end{tikzpicture}
\label{fig:hieR_eval_arch}}}
\caption{Performance of $\pi_c$ with different neural networks. GRU-Net and SA-Net are defined in Fig.~\ref{fig:network_architecture}.} 
\label{fig:hieR_results_arch}
\end{figure}

Next, we explore architectural modifications in the neural network used to train the commander. In the base training setup, we employ the GRU-Net as detailed in Sec.~\ref{sec:overall_structure}. We compare this with the SA-Net, which is used to train $\pi_f$, and also consider the FC-Net as discussed in Sec.~\ref{sec:fight_architect_inspect}. The performance of all three networks is depicted in Fig.~\ref{fig:hieR_results_arch}. Although all three networks achieve comparable mean rewards upon the completion of training, the evaluation clearly demonstrates that the GRU-Net delivers superior performance. Despite the self-attention module's capacity to focus on pertinent data within the observations, the GRU module appears more proficient in mastering strategic planning by incorporating the last states in the observations.

\subsubsection{Different Combat Scenarios}\label{sec:hieR_combat_scenes}
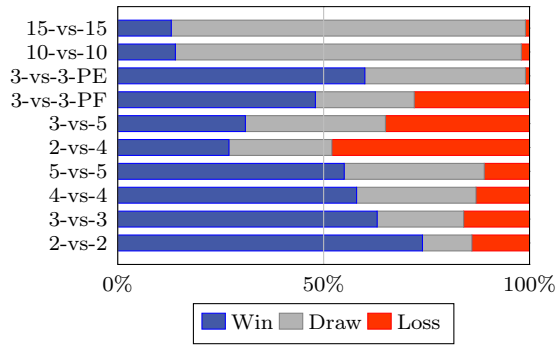
\begin{figure}[htb]
\centering
\begin{tikzpicture}
\begin{axis}[
xbar stacked,
legend entries={{\scriptsize Win},{\scriptsize Draw},{\scriptsize Loss}},
bar width=6,
width=7cm,
xmajorgrids,
height=5cm,
xmin=0,
xmax=100,
ytick={0,1,2,3,4,5,6,7,8,9},
yticklabels={{\footnotesize 2-vs-2},{\footnotesize 3-vs-3}, {\footnotesize 4-vs-4}, {\footnotesize 5-vs-5}, {\footnotesize 2-vs-4}, {\footnotesize 3-vs-5}, {\footnotesize 3-vs-3-PF}, {\footnotesize 3-vs-3-PE}, {\footnotesize 10-vs-10}, {\footnotesize 15-vs-15}},
xtick={0,50,100},
tickwidth=0mm,
legend style={at={(0.5,-0.15)},anchor=north,legend columns=-1},
xticklabels={{\scriptsize $0\%$},{\scriptsize $50\%$},{\scriptsize $100\%$}},
axis on top]
\addplot [color=blue, fill=blue!65!lime] coordinates 
{(74,0) (63,1) (58,2) (55,3) (27,4) (31,5) (48,6) (60,7) (14,8) (13,9)};
\addplot [color=gray, fill=gray!60!white] coordinates 
{(12,0) (21,1) (29,2) (34,3) (25,4) (34,5) (24,6) (39,7) (84,8) (86,9)};
\addplot [color=red, fill=red!60!orange] coordinates 
{(14,0) (16,1) (13,2) (11,3) (48,4) (35,5) (28,6) (1,7) (2,8) (1,9)};
\end{axis}
\end{tikzpicture}
\caption{Performance of our hierarchical MARL model under different combat scenarios and opponent behaviors. For opponents, PF denotes for \emph{Pure-Fight} ($p_o=\qty{100}{\percent}$) and PE denotes \emph{Pure-Escape} ($p_o=\qty{0}{\percent}$).}
\label{plot:best_hieR_diff_scenarios}
\end{figure}

As validated in Table~\ref{table:hieR_configs}, the CTDE approach \emph{Shared-N2-Opt-Assess}, as defined in Sec.~\ref{sec:commandeR_policy}, demonstrated the best performance for strategic decision-making. We further explore the strength of our hierarchical approach across various combat scenarios, with $\pi_c$ being trained in a pure 3-vs-3 setting. We extended the analysis by adjusting the number of entities, ranging from 2-vs-2 to 15-vs-15, and by varying the tactical behaviors of the opponents. The evaluation results are shown in Fig.~\ref{plot:best_hieR_diff_scenarios}. In smaller team configurations from 2-vs-2 to 5-vs-5, our commander policy consistently achieves a victory rate exceeding \qty{50}{\percent}, with a noted increase in draws as the number of participants rises. In uneven setups, where more opponents than agents are involved (2-vs-4 and 3-vs-5), the distribution of wins, losses, and draws is relatively balanced, demonstrating the robustness of our method even when engaging as a combat minority. When altering the opponents' combat tactics, distinct outcomes can be observed. In scenarios where opponents are exclusively set to fight mode (\emph{Pure-Fight}, PF), the commander policy still managed to achieve a win rate near \qty{50}{\percent}, albeit with a higher incidence of losses compared to the standard 3-vs-3 setting. Conversely, there is a minimal loss rate in scenarios where opponents attempt to escape (\emph{Pure-Escape}, PE). Despite an increased rate of draws in this scenario, the win rate remained predominant, further highlighting the superiority of our approach. We note that, $\pi_e$ includes firing actions, but was not rewarded during training for destroying an enemy. For larger configurations involving 10-vs-10 and 15-vs-15 entities, we extended the time horizon from $T=\num{500}$ to $T=\num{1000}$. The majority of these engagements resulted in draws, attributable to the higher likelihood of an agent or opponent surviving until the end of an episode due to the increased number of aircraft. Nonetheless, the favorable win-loss ratio underscores that our approach can effectively navigate and succeed in more challenging air combat scenarios.

\subsubsection{Commander-vs-Standard RL}\label{sec:hieR_vs_standard}
In the final phase of our study, we compare the performance of our hierarchical model against a conventional RL technique. This standard RL approach with policy $\pi_{\mathrm{std}}$ resembles the CTCE configuration delineated in Sec.~\ref{sec:hieR_architect_inspect}. We introduce two principal modifications to the previously shown CTCE framework. First, omission of curriculum learning, opting instead for direct training at L3 to facilitate optimal comparison due to the relatively deterministic opponent strategies. Second, integration of fight and escape rewards to simulate both behaviors, which in our method are distinctly defined through separate policies, $\pi_f$ and $\pi_e$. The total reward for training $\pi_{\mathrm{std}}$ is defined as:
\begin{equation}
    R_{\mathrm{std}} = R_k + R_d + R_{fr} + R_b + R_{p,t}\,.
\label{eq:standard_rl_rew}
\end{equation}

\begin{figure}[htb]
\centering
\subfloat[Training comparison $\pi_{\mathrm{std}}$-vs-$\pi_f$ (with curriculum stages).]{%
\resizebox*{6cm}{!}{\includegraphics{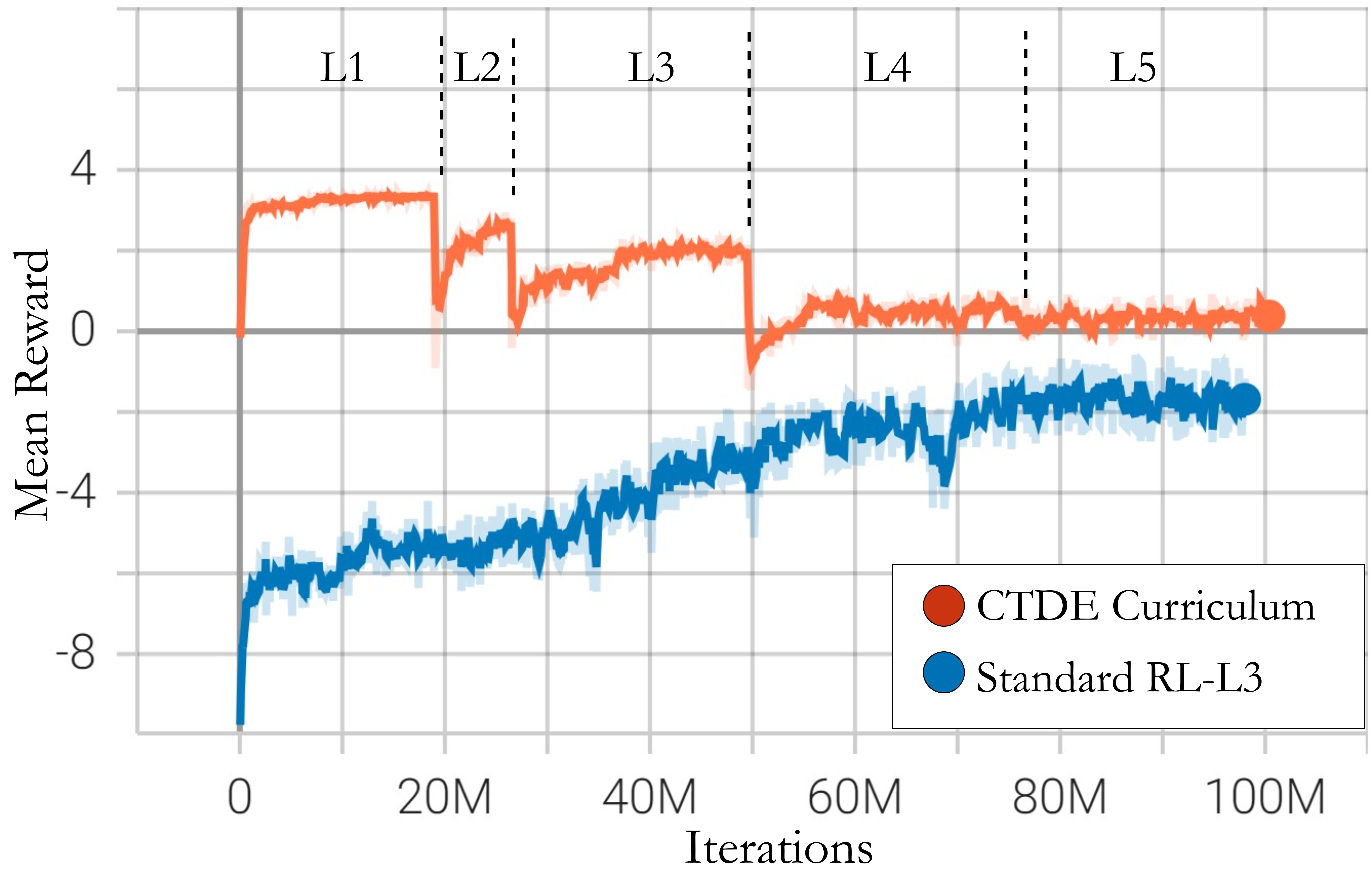}
\label{fig:standard_ctde_train}}}
\hspace{5pt}
\subfloat[Evaluation comparison $\pi_{\mathrm{std}}$-vs-$\pi_c$ with $\pi_{f,L5}$ and $\pi_e$.]{%
\resizebox*{6cm}{!}{
\begin{tikzpicture}
\begin{axis}[
xbar stacked,
legend entries={{\scriptsize Win},{\scriptsize Draw},{\scriptsize Loss}},
bar width=10,
width=5cm,
xmajorgrids,
height=2.6cm,
xmin=0,
xmax=100,
ytick={0,1},
yticklabels={{\footnotesize Standard RL}, {\footnotesize H-MARL}},
xtick={0,50,100},
tickwidth=0mm,
legend style={at={(0.5,-0.6)},anchor=north,legend columns=-1},
xticklabels={{\scriptsize $0\%$},{\scriptsize $50\%$},{\scriptsize $100\%$}},
axis on top]
\addplot [color=blue, fill=blue!65!lime] coordinates
{(16,0) (63,1)};
\addplot [color=gray, fill=gray!60!white] coordinates
{(51,0) (21,1)};
\addplot [color=red, fill=red!60!orange] coordinates
{(33,0) (16,1)};
\end{axis}
\end{tikzpicture}
\label{plot:standard_hieR_eval}}}
\caption{Performance comparison of Standard RL-vs-Hierarchical MARL (H-MARL) approach.} 
\label{fig:standard_hieR_compare}
\end{figure}
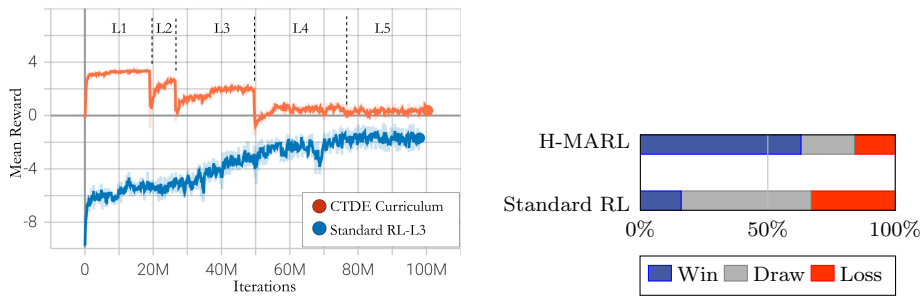

It is noteworthy that $R_{p,t}$ from Eq.~\eqref{eq:esc_p} is adjusted to avoid penalizing the agent for proximity to opponents, specifically, $R_{p,t} = +\num{0.1}$ if $d>\qty{13}{\kilo\metre}$ is given per-time-step. Thus, policy $\pi_{\mathrm{std}}$ is required to autonomously discern when evasive actions supersede attacking maneuvers. For simplification in training $\pi_{\mathrm{std}}$, we keep the configuration of low-level policies by sensing only the nearest opponent.
In Fig.~\ref{fig:standard_ctde_train}, we evaluate the training performance of $\pi_{\mathrm{std}}$ relative to our entire curriculum training sequence for the low-level policy $\pi_f$, with the specific levels indicated within the graph. Notably, $\pi_{\mathrm{std}}$ fails to achieve the performance that $\pi_f$ reaches at only half the total training iterations, specifically at the end of L3 in $\pi_f$ training. Subsequently, we assess the 3-vs-3 performance of $\pi_{\mathrm{std}}$ against the commander policy $\pi_c$. The evaluation results are shown in Fig.~\ref{plot:standard_hieR_eval}. The hierarchical outcomes reflect those observed in the 3-vs-3 evaluation scenario presented in Fig.~\ref{plot:best_hieR_diff_scenarios}. The standard RL strategy employing $\pi_{\mathrm{std}}$ does not match the superior performance of the commander policy, that incorporates $\pi_{f,L5}$ and $\pi_e$ for distinct behaviors. This comparison underscores the strategic and operational advantages attributable to our design choices in the context of tactical air combat maneuvering. An example of a 2-vs-4 combat scenario with the commander involved is shown in Fig.~\ref{fig:traj_hier}, where two opponents could successfully be destroyed within the time horizon.

\section{Conclusion}\label{sec:conclusions}
\subsection{Review}\label{sec:conclusion_review}
We introduced a heterogeneous hierachical MARL approach to investigate preset air combat simulations. Our methodology combines several recent developments in the MARL area, including the usage of a learning curriculum, fictitious self-play techniques, and hierachy-specific neural networks. Empirical validation demonstrates the promising potential of our approach. This validation involved a detailed ablation study, where we measured the effect of modifying principal system components on the overall combat performance. Our agents exhibit effective air combat capabilities across diverse combat scenarios, even when scaled to large team configurations. We find that trained low-level agents are able to execute effective decisions autonomously but, at the same time, the employment of a high-level commander improves the overall performance by look-ahead planning. Curriculum-based training has emerged as an effective design strategy for training of low-level agents. By providing tactical commands for each agent separately, the simple yet effective training setup of the commander achieved reasonable performance and offers sufficient flexibility for various combat configurations. We observed that rewarding agents solely based on their combat achievements yields better overall performance than a reward sharing mechanism. For the commander agent, the introduction of intrinsic reward signals led to a positive performance impact. The SA-module within our neural network architecture resulted in a significant improvement, and for the commander, the GRU-module has taken a pivotal role. In our analysis, we noted that reduced sensing capability (N2) leads to better combat tactics, even though it provides the commander with fewer options for decision-making. Finally, the CTDE framework has proven to be effective in the coordinated training of agents, showing its power even in scenarios with heterogeneous agents.

\subsection{Future Work}\label{sec:conclusion_future}
In detailed inspections of some rendered combat scenrios we noticed that fleeing maneuvers occur even under the fight policy $\pi_f$, e.g. in scenarios with (significantly) more alies than opponents. This might be viewed as a fail case since rather than leaving the environment boundaries, agents should continue their engagements and tactical behavior should exclusively be provided by the commander agent. This behavior may be attributable to the insufficient training of individual agents in highly asymmetric scenarios. It can be expected that, more generally, edge-case scenarios reveal sub-optimal decision-making. In future research, we aim to enhance the diversity of low-level policies for a more fine-grained behavioral distinction. This includes in particular the investigation of detailed commands from the high-level policy, beyond the binary attack/evade choice described in this work. Potential commands could include specific situational information such as \lq\lq{}aggressive attacking\rq{}\rq{}, \lq\lq{}defending the position\rq{}\rq{}, the choice of weapon and similar.
We aim to improve the training of the commander policy by using dedicated planning algorithms like Monte-Carlo Tree Search or AlphaZero~\cite{alphaZero}. This might include combat scenarios in the BVR setting with different weapons and sensing strategies. 
We are in the process of developing a 3D simulation environment based on the JSBSim package for precise aircraft dynamics\footnote{ \href{https://jsbsim.sourceforge.net/}{https://jsbsim.net}}, thereby making a step towards realistic air combat simulations.



\bibliographystyle{files/tfnlm}
\bibliography{2biblio}

\newpage

\appendix

\end{document}